\documentclass[conference]{IEEEtran}
\usepackage{times}
\usepackage[numbers]{natbib}
\usepackage[bookmarks=true]{hyperref}
\usepackage{multicol}
\usepackage{subcaption}

\usepackage{caption}
\usepackage{listings}
\usepackage{graphicx}
\usepackage{amsmath}
\usepackage{amsthm}
\usepackage{xcolor}
\usepackage{amsfonts}
\usepackage{multirow}
\usepackage{algorithm}
\usepackage{algpseudocode}
\usepackage{algorithmicx}
\usepackage{bbold}
\usepackage{caption}
\usepackage{mathtools}
\usepackage{booktabs}
\let\labelindent\relax
\usepackage{enumitem}

\usepackage{graphicx}
\usepackage{booktabs}
\usepackage{caption}
\usepackage{algorithm}
\usepackage[table]{xcolor} 
\usepackage{tcolorbox}
\tcbuselibrary{breakable}
\usepackage{tikz}
\usetikzlibrary{positioning, arrows.meta, shapes.geometric}

\usepackage{tikz}

\usepackage{amsmath,amsfonts,bm}
\usepackage{xspace}

\def\figref#1{Fig.~\ref{#1}}

\def\Secref#1{Section~\ref{#1}}

\def\eqref#1{equation~\ref{#1}}

\def\1{\bm{1}}

\DeclareMathAlphabet{\mathsfit}{\encodingdefault}{\sfdefault}{m}{sl}
\SetMathAlphabet{\mathsfit}{bold}{\encodingdefault}{\sfdefault}{bx}{n}

\renewcommand{\eqref}[1]{(\ref{#1})}

\newcommand{\tabref}[1]{Table~\ref{#1}}

\newcommand{\observation}{\ensuremath{o}}

\newcommand{\tightsection}[1]{\vspace{-1pt}\section{#1}\vspace{-1pt}}
\newcommand{\tightsubsection}[1]{\vspace{-1pt}\subsection{#1}\vspace{-1pt}}

\usepackage{xspace}
\newcommand{\redtext}[1]{\textcolor{black}{#1}}
\newcommand{\bluetext}[1]{\textcolor{black}{#1}}
\renewcommand{\eqref}[1]{(\ref{#1})}

\newcommand{\eg}{\textrm{e.g.}}

\theoremstyle{definition}  %

\newboolean{bcmt}
\setboolean{bcmt}{true}
\usepackage{ifthen}

\usepackage{color}
\definecolor{darkgreen}{rgb}{0,0.5,0}
\definecolor{fullred}{rgb}{0.95,.0,.1}
\definecolor{brown}{rgb}{0.65,0.16,0.16}
\definecolor{orange}{rgb}{1,0.5,0}
\definecolor{lightblue}{HTML}{F0F0F0}

\newcounter{prompt}

\usepackage{amsmath,amssymb}
{\begin{list}{$\bullet$}{%
    \setlength{\topsep}{0in}
    \setlength{\partopsep}{0in}
    \setlength{\itemsep}{0in}
    \setlength{\parsep}{0in}
    \setlength{\leftmargin}{3.5em}
    \setlength{\rightmargin}{0in}
    \setlength{\itemindent}{0in}
}
}%
{\end{list}}

\newcommand{\agenta}{\ensuremath{1}}
\newcommand{\agentb}{\ensuremath{2}}

\newcommand{\statespace}{\ensuremath{S}}

\newcommand{\Tdots}{\ensuremath{T}}
\newcommand{\Rdots}{\ensuremath{R}}

\newcommand{\Aactionspace}{\ensuremath{A^\agenta}}
\newcommand{\Bactionspace}{\ensuremath{A^\agentb}}

\newcommand{\name}{CANINE\xspace}

\newcommand{\skill}{\ensuremath{\omega}}
\newcommand{\skillset}{\ensuremath{\Omega}}

\begin{document}

\title{\name{}: Coaching Visually Impaired Users for Interactive Navigation with a Robot Guide Dog}

\author{
\authorblockN{Cunjun Yu$^{1}$,
 Zishuo Wang$^{1}$,
 Anxing Xiao$^{1}$, 
 Linfeng Li$^{1}$,
 David Hsu$^{1,2}$}
\authorblockA{
  $^1$School of Computing,
  $^2$Smart Systems Institute \\
  National University of Singapore \\
}}

\twocolumn[{%
\renewcommand\twocolumn[1][]{#1}%
\maketitle
}]

\IEEEpeerreviewmaketitle

\begin{abstract}
\textit{Robot} guide dogs offer navigation assistance that greatly expands the independent mobility of the visually impaired, but their effective use requires subtle  human-robot coordination that is difficult for  users to learn from generic verbal instructions. To tackle this challenge, we present \name{}, an automated coaching system that trains  users for \textit{interactive navigation} with a robot guide dog, through  personalized, adaptive verbal feedback. \name{} decomposes a complex coordination task into sub-skills and operates at two levels. At the high level, it decides \textit{what} to train by tracking the learner's proficiency across sub-skills using \textit{knowledge tracing} and  prioritizing training on the weakest areas. At the low level, \name{} decides \textit{how} to train  each sub-skill by observing each human practice episode, using foundation models to infer the underlying causes of errors, and generating targeted verbal corrections adaptively. \redtext{A controlled study with blindfolded participants, treated as a proxy population for quantitative evaluation, demonstrates that \name{} significantly improves both learning efficiency and final navigation performance compared to generic verbal instructions.} We further validate \name{} through  a retention study and  \redtext{an exploratory} case study.  The retention study shows lasting skill improvement after two weeks. The case study confirms \name{}'s effectiveness in training a visually impaired user, while revealing additional design considerations for real-world deployment. Both are well aligned with the findings of the controlled study. The video can be found on the \href{https://cunjunyu.github.io/project/canine/}{project page}.
\end{abstract}
  
\tightsection{Introduction}
\label{sec:intro}

\begin{figure}[t]
  \centering
  \includegraphics[width=\columnwidth]{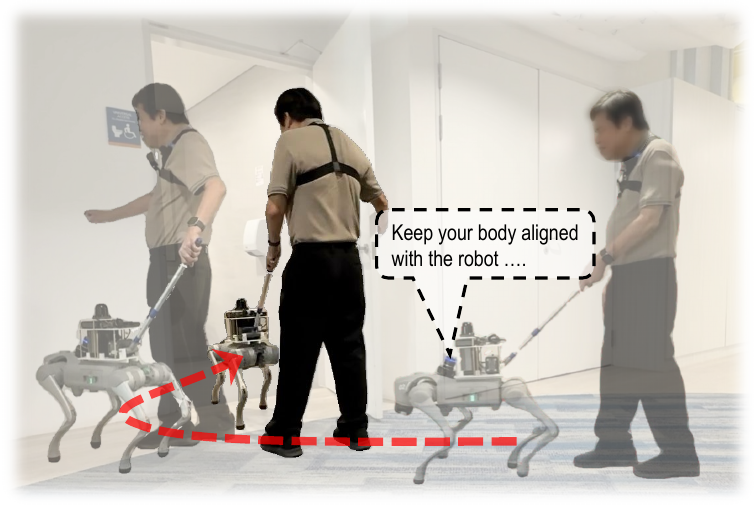}
  \caption{\textbf{CANINE}. In our study, a robot guide dog coaches a visually impaired user to navigate through a doorway.}
  \vspace{-20pt}
  \label{fig:teaser}
\end{figure}

For the 2.2 billion people worldwide with visual impairments (VI) according to the WHO~\cite{who2023vision}, everyday tasks like navigating unfamiliar environments require assistance that is often unavailable~\cite{MIZUKOSHI2008193}. While robotic guide systems promise to provide on-demand support~\cite{xiao2021robotic, chen2023quadruped, cai2024navigating, Hwang2026GuideNav, guerreiro2019cabot, kamikubo2025beyond,kuribayashi2023pathfinder}, a critical challenge remains largely unaddressed: \textit{how can we effectively coach users to leverage these systems?} Consider a user navigating through a doorway with a robot guide dog~\cite{Cai2026NavigationBeyondWayfinding} as shown in~\figref{fig:teaser}. This task requires precise spatial positioning relative to both the robot and the door frame, combined with exact coordination to avoid collisions, all without visual feedback. Similarly, when such robotic platforms are equipped with a manipulator arm for object handover~\cite{chan2013human, kupcsik2017learning}, users must coordinate hand positioning and grasp timing without seeing the robot's approach. Without effective coaching, users may develop unsafe habits, and experience repeated failures that erode confidence. These scenarios exemplify a broader class of physical interaction tasks where one agent (the human) lacks key perceptual channels and must learn to coordinate with another agent (the robot) through alternate modalities.

Coaching humans to master these tasks presents two fundamental challenges. First, these tasks are complex processes composed of distinct phases, each requiring different coordination strategies. For example, locating a door handle requires spatial exploration, whereas passing through the door requires precise coordination. Given the heterogeneity of these phases and significant variability across users, treating the task as a single monolithic process is inefficient. Second, the user's lack of visual perception makes providing feedback challenging. The user cannot observe the environment to correct their errors and must rely on the coaching system to provide in-situ feedback. To provide meaningful guidance, the coach must infer latent user states, such as distinguishing between spatial misalignment and cue misinterpretation, solely from noisy interaction signals. Existing approaches rely on periodic sessions with orientation and mobility (O\&M) specialists~\cite{om2012}, which are effective but require human professionals and provide limited opportunities for repetitive skill refinement. 

We present \name{}, an automated coaching system that trains users for interactive navigation with robot guide dogs through personalized, adaptive verbal feedback. \name{} addresses these challenges by operating at two distinct levels, mirroring how human coaches determine \textit{what} to practice and \textit{how} to practice it. To enable targeted training, \name{} decomposes the overall task into \textit{sub-skills}, modular components that can be practiced independently, \eg{}, open door. 

At the \textit{inter-skill level}, \name{} addresses curriculum scheduling by tracking the learner's latent proficiency across sub-skills using \textit{knowledge tracing}~\cite{Wilson2016BackTT,knowledgetracingKnowledge} and dynamically prioritizing training in the weakest areas. We formulate this as a partially observable decision problem where user proficiency cannot be directly observed but must be inferred from performance, enabling the system to maintain beliefs about user competencies and select training activities that maximize learning efficiency. At the \textit{intra-skill level}, \name{} addresses feedback generation by observing each practice episode and providing adaptive audio feedback through a speaker at the end of each attempt. The system observes the user's behavior, uses foundation models to infer the underlying causes of errors (e.g., whether an error stems from spatial misalignment or misunderstanding of robot cues), and generates targeted verbal corrections. To handle the complexity of modeling human behavior, we instantiate an approximate policy using a structured reasoning process: Vision-Language Models (VLMs)~\cite{openai2023gpt4} extract behavioral states from video observations, and Large Language Models (LLMs)~\cite{brown2020language} generate actionable feedback tailored to the inferred state (e.g., ``position yourself further from the robot'').

We validate \name{} through a multi-stage evaluation on doorway navigation with robot guide dogs. \redtext{In a controlled proxy-user study with blindfolded sighted participants~\cite{xiao2021robotic, chen2023quadruped}, \name{} significantly improves learning efficiency and final performance over generic verbal instructions.} We further validate \name{} through (i) a retention study showing lasting skill improvement after two weeks and (ii) \redtext{an exploratory} case study with a visually impaired user.

\begin{figure*}[t]
  \centering
  \hspace{-3pt}\includegraphics[width=0.97\textwidth]{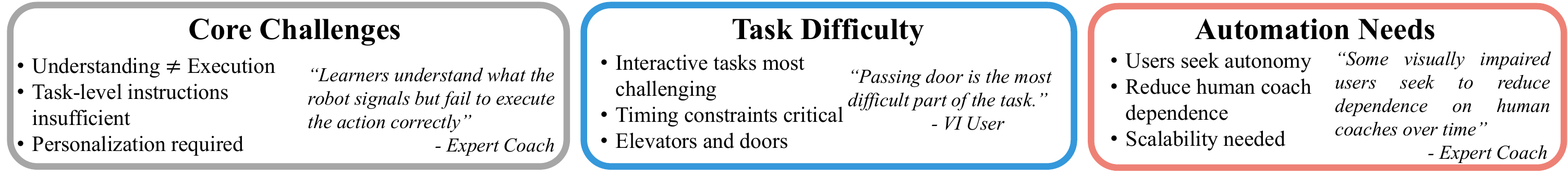}
  \caption{\textbf{Expert coach insights from formative study.} Analysis reveals common learner challenges, effective coaching strategies (timing and content), and design implications for automated coaching systems.}
  \label{fig:expert_insights}
  \vspace{-15pt}
\end{figure*}

\tightsection{Related Works}
\label{sec:related}
Our work sits at the intersection of assistive robotics, human-robot interaction, and intelligent tutoring systems.

\noindent\textbf{Physical Human-Robot Interaction.}
Physical Human-Robot Interaction research has explored three paradigms: \emph{assistance}, \emph{collaboration}, and \emph{teaching}~\citep{6161710, 9501975}. Assistance systems infer human intent and adapt robot behavior to help users achieve immediate goals~\citep{Dragan2013,ReddyDL18,wu2025savor,jenamani2025feast,padmanabha2026waffle}, such as mobility augmentation~\citep{molinaro2024task} or motor function restoration~\citep{chen2024upper}. Collaboration systems enable humans and robots to work toward shared objectives through mutual adaptation~\citep{li2015continuous}, as seen in object manipulation~\citep{Nikolaidis2016} and co-manipulation tasks~\citep{rozo2016learning}. Teaching systems structure interactions to improve human capabilities over time~\citep{yu2022coach,Tian2023TowardsMA,li2024towards,Qian2025AS,gopinath2024computational,srivastava2025shared}. Recent robotic teaching work has made significant progress on key aspects of the coaching problem. Curriculum-focused methods~\citep{yu2022coach,srivastava2025shared,srivastava2022assistive,du_grannen2026snoopie} have demonstrated effective approaches enabling adaptive skill sequencing. Feedback-focused methods~\citep{Qian2025AS,srivastava2023generating,lynn2026skill} have shown how to monitor performance and generate corrective guidance. Physical assistance methods~\citep{houyu2026ral} have advanced techniques for modulating robot behavior during execution to support motor learning. Building on these foundations, \name{} integrates proficiency-based curriculum scheduling with diagnostic verbal feedback to address the complete coaching loop, particularly for learners who cannot visually self-correct.

Assistive robotics has developed navigation systems for visually impaired users~\citep{xiao2021robotic, chen2023quadruped, cai2024navigating}. While these systems demonstrate technical feasibility in controlled settings, they provide no structured learning support, leaving users to acquire coordination skills through trial and error. \name{} addresses this gap by providing systematic coaching.

\noindent\textbf{Intelligent Tutoring Systems.}
Intelligent Tutoring Systems (ITS) have a long history in education, focusing on personalized instruction and adaptive learning~\citep{psotka1988intelligent,irt,Clement_Roy_Oudeyer_Lopes_2015, knowledgetracingKnowledge}. However, traditional ITS primarily target cognitive skills (e.g., mathematics, language learning)~\citep{swartz2012intelligent,lu2021radarmath, steenbergen2013meta} rather than embodied, physical skills. Recent work has begun to explore LLM-based coaching for physical activities, such as GPTCoach~\cite{gptcoach} which provides feedback based on self-reported performance data and sensor readings. GPTCoach takes in user self-reported structured performance metrics (e.g., exercise counts) and generates motivational feedback and workout plans. In contrast, \name{} must diagnose coordination errors from raw visual observations of human-robot interaction, requiring the system to infer what went wrong from unstructured observation alone and generate actionable corrections for spatial and temporal coordination failures that users cannot directly observe.

\noindent\textbf{Foundation Models for Robotics.}
Recent advances in foundation models, such as LLMs and VLMs, have shown promise for robotics applications~\citep{firoozi2025foundation, kim2024understanding,zitkovich2023rt2,liang2023code,driess2023palme, ahn2022saycan, shah2023vint,team2024octo, black2025pi0,irfan2025between, wang2024lami}. These models have been used for task planning, natural language instruction following, and scene understanding. However, most applications focus on enabling robots to execute tasks more effectively using language for task specification and clarification~\cite{shi2025hi, cui2023no}, rather than using these models to teach humans. \textrm{In contrast to prior work that uses foundation models to improve robot task execution, \name{} inverts this paradigm by using VLMs and LLMs as reasoning tools to coach humans while robots provide physical interaction.}

\tightsection{Coaching Needs and Design Principles}
\label{sec:formative}
To understand the coaching needs of VI users learning to use robot guide dogs, we conducted a formative study with an expert coach (2.5 years experience with robotic guiding technology) and one VI participant (2 years experience with robot guide dogs). \figref{fig:expert_insights} summarizes key results.

\redtext{The formative study used a semi-structured interview built around a questionnaire organized into six sections: background and prior assistive-technology training; robot-guide-dog experience; coaching and feedback preferences; learning preferences; safety and trust; and future use. We analyzed responses for recurring themes, including limited opportunities for repeated expert coaching and a preference for error-specific, actionable feedback. These themes directly informed three design choices: decomposing practice into sub-skills, generating personalized terminal feedback, and adapting the training sequence to each learner's observed weaknesses. The full instrument is provided in Appendix~\ref{appendix:formative_questionnaire}.}

\noindent\textbf{The core challenge: understanding $\neq$ execution.} General instructions are insufficient because learners often understand the high-level goals but fail to execute the corresponding physical actions, leading to diverse error patterns. Effective coaching must therefore go beyond explaining ``what'' to do and adaptively diagnose ``how'' the execution failed, modifying feedback to correct specific physical errors.

\noindent\textbf{Interactive tasks with timing constraints are most difficult.} Tasks requiring active physical coordination, such as navigating doorways, particularly under time pressure, were identified as the most challenging. We thus select doorway navigation as our focal task.

\noindent\textbf{Automated coaching enables autonomy and scalability.} Both participants confirmed the necessity of dedicated coaching sessions for effective learning. In addition, visually impaired users seek autonomy and to reduce dependence on human coaches over time. While structured coaching is essential for effective learning, automating adaptive coaching would scale the process and provide independence to practice without constant supervision.

\tightsection{Problem Formulation}
\label{sec:formulation}
We formulate robotic assistance coaching as a two-part problem: a \textit{target task} where humans learn to coordinate with robotic assistants, and a \textit{teaching task} where a robot coach optimizes the training process.

\noindent\textbf{Target Task: Collaborative Navigation.}
We model the target task as a cooperative Markov Game $\mathcal{M}=( \statespace, \Aactionspace, \Bactionspace, \Tdots, \Rdots, \gamma )$. For doorway navigation, the state $\statespace$ encodes the joint physical configuration: human pose relative to the robot and door, robot pose, door state (closed/open/opening), etc. Actions comprise continuous human control inputs $\Aactionspace$ (body velocity, hand forces on door handle) and robot actions $\Bactionspace$ (base velocity). Transitions $\Tdots$ govern the coupled non-linear dynamics, requiring tight coordination (e.g., maintaining leash contact while opening the door). The cooperative reward $\Rdots$ incentivizes safe, efficient task completion (reaching the goal beyond the doorway) while penalizing collisions and loss of connection. The objective is for the human to learn a policy $\pi^H$ that complements the fixed robot policy $\pi^R$ to maximize expected return.

\noindent\textbf{Teaching Task: Hierarchical Coaching.}
We decompose the target task into discrete sub-skills $\skillset$ (e.g., for doorway navigation: Navigate to Door, Open Door, and Enter Room) and formulate coaching as two POMDPs:

\paragraph{Inter-skill POMDP: Curriculum Sequencing}
At the episode level, the coach solves $\mathcal{M}_{inter} = (\mathcal{S}_{skill}, \mathcal{A}_{skill}, \mathcal{O}_{perf}, \mathcal{T}_{skill}, \mathcal{R}_{learning})$. The state $\mathcal{S}_{skill}$ represents the learner's latent proficiency across sub-skills (a vector of continuous proficiency values, one per sub-skill). Actions $\mathcal{A}_{skill} = \skillset$ select the next sub-skill to practice. Observations $\mathcal{O}_{perf}$ are noisy performance metrics for the practiced sub-skill (e.g., completion time, success/failure). Transitions $\mathcal{T}_{skill}$ model how proficiency evolves with practice. The reward $\mathcal{R}_{learning}$ incentivizes practicing sub-skills where the learner has low proficiency, thereby maximizing overall learning progress. The coach cannot directly observe proficiency and must maintain a belief over $\mathcal{S}_{skill}$ based on observed performance.

\paragraph{Intra-skill POMDP: Feedback Generation}
Within a chosen sub-skill $\omega$, the coach solves $\mathcal{M}_{intra}^\omega = (\mathcal{S}_{task}, \mathcal{A}_{coach}, \mathcal{O}_{intra}, \mathcal{T}_{task}, \mathcal{R}_{guide})$. The state $\mathcal{S}_{task}$ captures the physical system state during task execution (human pose, robot pose, environment state, task progress) as well as latent cognitive states (whether errors stem from spatial misalignment or cue misinterpretation). Actions $\mathcal{A}_{coach}$ consist of post-episode verbal feedback and robot parameter adjustments for the next attempt. Observations $\mathcal{O}_{intra}$ are visual observations from a chest-mounted camera plus optional user verbal feedback. \redtext{Optional verbal feedback is captured through a microphone, transcribed to text, and passed to the coaching model together with the visual episode summary.} The reward $\mathcal{R}_{guide}$ incentivizes feedback that corrects errors and ensures safety. The coach must infer the latent cognitive state from observations to provide guidance.

Exact solutions are intractable due to continuous state spaces, latent human cognition, and unknown learning dynamics. We therefore employ approximate solutions tailored to each level: Bayesian knowledge tracing with greedy action selection for the inter-skill POMDP, and a structured VLM/LLM process for the intra-skill POMDP. 

\tightsection{\name{}}
\label{sec:method}
\begin{figure*}[!t]
  \centering
  \includegraphics[width=0.98\textwidth]{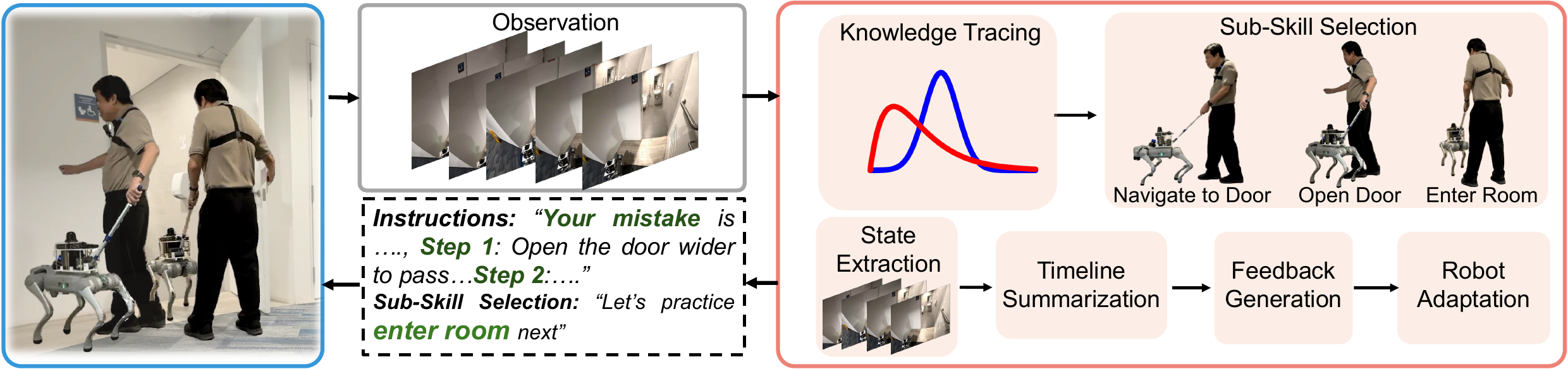}
  \caption{\textbf{Overview of \name{}.} \name{} employs a two-level coaching strategy. The \textbf{inter-skill} coaching (up-right) tracks proficiency across sub-skills using knowledge tracing and selects the sub-skill to practice next. The \textbf{intra-skill} coaching (down-right) takes in video observations and generates coaching instructions for the selected sub-skill. }
  \label{fig:overview}
    \vspace{-15pt}
\end{figure*}

\label{subsec:algorithm}
\tightsubsection{Overview}

\name{} is a two-level coaching system that enables visually impaired users to learn effective coordination with robot guide dogs. We decompose the overall navigation task into discrete sub-skills (e.g., for the doorway navigation task: Navigate to Door, Open Door, and Enter Room) that can be practiced independently. As the user navigates with the robot, the system observes behavior through a chest-mounted camera capturing a first-person view. The \emph{inter-skill} module dynamically sequences the curriculum by estimating proficiency across sub-skills and selecting which sub-skill to practice next. The \emph{intra-skill} module generates personalized feedback for each practice episode, leveraging foundation models to diagnose specific errors (e.g., distinguishing spatial misalignment from cue misinterpretation), provide corrections, and adjust robot behaviors. To minimize cognitive load during safety-critical maneuvers, feedback is delivered after each episode completes.

\tightsubsection{Inter-skill Coaching}
\label{subsec:inter_skill}
\redtext{The inter-skill layer uses the POMDP formulation for adaptive curriculum selection: the learner's proficiency across sub-skills is latent, each practice episode provides a noisy performance observation, and the coach chooses the next sub-skill to practice based on the current belief.} 

\noindent\textbf{Sub-skill Decomposition.} For doorway navigation, we decompose the task into three sub-skills: (1) Navigate to Door, (2) Open Door, and (3) Enter Room. Each sub-skill $\skill$ represents a coherent behavior that can be practiced and evaluated independently by configuring the robot to start from the appropriate initial state (e.g., for ``Open Door'', the robot positions the user in front of a closed door).

\noindent\textbf{Knowledge Tracing.} The learner's true proficiency for each sub-skill is latent and cannot be directly observed. The coach maintains a belief $b^t$ over proficiency, modeling each skill with a Gaussian distribution $\mathcal{N}(\mu^t_\skill, (\sigma^t_\skill)^2)$, extending Bayesian knowledge tracing~\citep{Wilson2016BackTT, knowledgetracingKnowledge} to continuous performance metrics.

When the learner practices sub-skill $\skill$, the coach observes a performance measurement $y_t$ (e.g., completion time). The coach maintains a Gaussian belief over the learner’s latent expected performance,
$p(\mu_\skill^t) = \mathcal N(\mu_\skill^t, (\sigma_\skill^t)^2)$.
To model gradual skill drift, the prior variance is first inflated by adding process noise,
$
(\sigma_\skill^t)^2 \leftarrow (\sigma_\skill^t)^2 + \sigma_{\text{proc}}^2.
$
The observation $y_t$ is then incorporated via a Bayesian update under a linear-Gaussian observation model. The posterior mean $\mu_\skill^t$ is mapped to a normalized proficiency score $p_\skill^t \in [0,1]$ by linearly interpolating between novice and expert baseline performance,
$
p_\skill^t
= \mathrm{clip}\!\left(
\frac{\mu_{\text{novice}} - \mu_\skill^t}
     {\mu_{\text{novice}} - \mu_{\text{expert}}},
\, 0, 1 \right),
$
where $\mu_{\text{expert}}$ and $\mu_{\text{novice}}$ denote expected completion times for expert and novice performance, respectively. This mapping assigns proficiency $1.0$ to expert-level performance, $0.0$ to novice-level performance, with linear interpolation between.

\noindent\textbf{Curriculum Policy.} At each step, the system selects the sub-skill with the lowest estimated proficiency for practice. This greedy policy prioritizes the learner's weakest areas, approximating the optimal curriculum under the assumption that practicing weak skills yields the highest learning gain. \redtext{This gives a concrete curriculum policy for the inter-skill decision process: the belief tracker estimates which proficiency state is weakest, and the selected sub-skill becomes the next coaching action.} The curriculum adapts dynamically as proficiency estimates evolve. \redtext{Initial proficiency beliefs are personalized using each participant's initial practice trials: the first observations initialize the participant-specific belief means, while expert and novice reference performance from a pilot study set the normalization endpoints and tracker hyperparameters. Thus, the pilot data calibrates the scale of the tracker, but each participant's curriculum is initialized from their own measured performance in initial practice trials.}

\tightsubsection{Intra-skill Coaching}
\redtext{Within a selected sub-skill, the coach faces the partial-observation problem described in \Secref{sec:formulation}: the chest-mounted camera reveals the user's behavior and task progress, but not the underlying cause of an error.} The coach must infer the latent cognitive states from observations to provide appropriate guidance. \redtext{Accordingly, our intra-skill pipeline converts visual observations into symbolic state histories, summarizes the episode, diagnoses likely latent error causes, and then produces both verbal feedback and robot-parameter adjustments.} For the doorway navigation task in this paper, we use terminal (post-episode) feedback to avoid interfering with time-critical safety behaviors during navigation.

Let $\observation_{1:T}$ denote the raw visual observation sequence (RGB images) from a chest-mounted camera during a practice episode. We use a structured four-stage process that processes visual observations, infers user states, and generates coaching actions:
\begin{equation}
    \observation_{1:T}
    \xrightarrow{\;f_{\text{frame}}\;}
    h_{1:T}
    \xrightarrow{\;f_{\text{time}}\;}
    \tau
    \xrightarrow{\;f_{\text{coach}}\;}
    c
    \xrightarrow{\;f_{\text{param}}\;}
    u,
    \label{eq:intra_pipeline}
\end{equation}
where $h_t$ is the symbolic state representation at time $t$, $\tau$ is the episode summary, $c$ is the coaching action (verbal instructions), and $u$ are robot parameter adjustments to alter the behavior of the robot. \redtext{The intermediate history $h_{1:T}$ and summary $\tau$ form the coach's episode-level belief context, while $c$ and $u$ are the resulting coaching actions exposed to the user and robot.} These components execute once per episode (after time $T$), ensuring the robot's real-time control depends only on its low-level controller, not on LLM latency.
\redtext{The per-stage I/O details and prompts can be found in Appendix.}

\noindent\textbf{Frame-level State Extraction ($f_{\text{frame}}$).} This module takes as input the raw visual observation sequence $\observation_{1:T}$ and produces symbolic state representations $h_{1:T}$. Given the observation trajectory, we uniformly sample key frames and apply $h_t = f_{\text{frame}}(\observation_{1:T}, t)$ to each timestamp $t$. In practice, we capture an image every 0.5 seconds and feed it to the VLM. The output $h_t$ encodes: user physical state (approaching, positioned, interacting), robot state, environment state (clear, cluttered, hazardous), task progress (not started, in progress, complete, failed), and safety flags (collision risk, blocked path). This structured representation extracts key information from the raw input to allow the LLM to reason about the user's behavior explicitly using symbolic representation.

\begin{figure}[!t]
    \centering
    \includegraphics[width=\columnwidth]{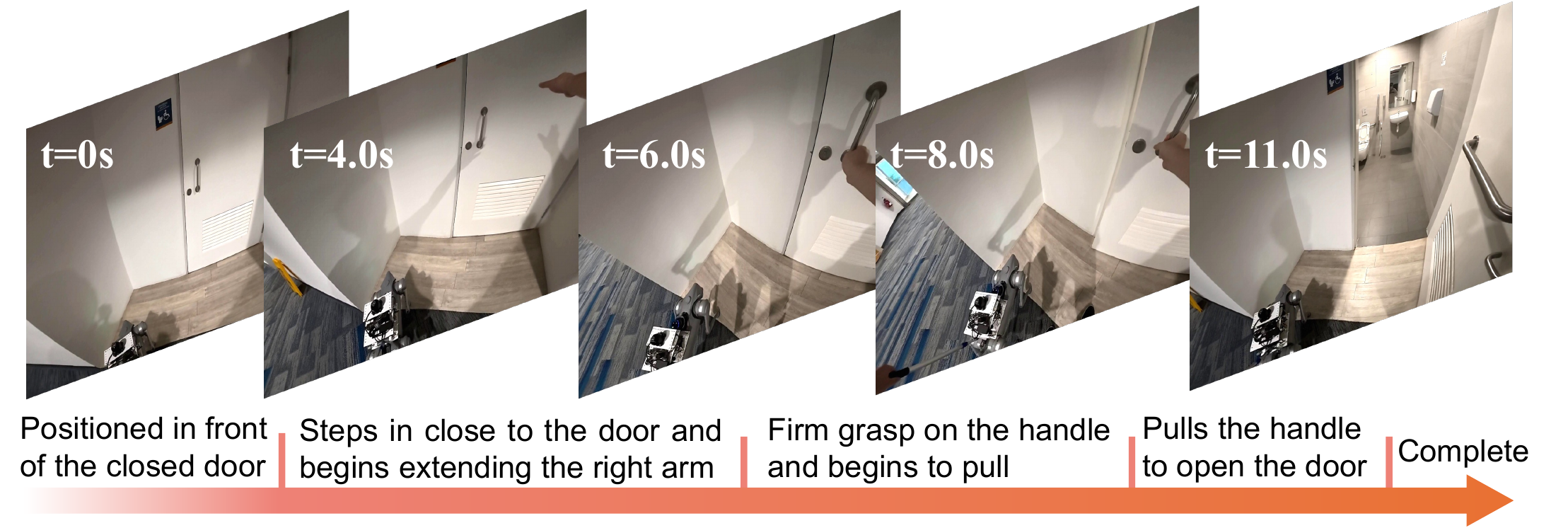}
    \caption{\textbf{Illustration of the summarized timeline.} The timeline aggregates per-frame observations sampled every 0.5\,s into a structured episode-level summary.}
    \label{fig:timeline}
    \vspace{-20pt}
\end{figure}

\noindent\textbf{Timeline Summarization ($f_{\text{time}}$).} This module takes as input the frame-level states $h_{1:T}$ and produces a compact episode summary $\tau$. While the frame sequence captures observable physical states, latent cognitive components (error causes, mental model) remain hidden. Applying $\tau = f_{\text{time}}(h_{1:T})$ produces a structured summary containing: (1) episode statistics (duration, terminal state, success/failure), (2) performance metrics (efficiency, safety scores), and (3) a discrete event sequence with timestamps (task start, progress milestones, failures, safety violations). One example can be found in~\figref{fig:timeline}. This makes behavioral patterns such as hesitation, repeated errors, or unsafe behaviors explicit and observable.

\noindent\textbf{Feedback Generation ($f_{\text{coach}}$).} This module takes as input the episode summary $\tau$ and optional user verbal feedback, and produces a coaching action $c$ containing an error diagnosis and a verbal instruction. We use an LLM to infer the likely latent cognitive state (i.e., the error cause) from the observable evidence: it analyzes the summary and user feedback to diagnose the source of errors (e.g., spatial misalignment, cue misinterpretation), then generates the coaching action. \bluetext{User feedback helps disambiguate the visual summary, and the diagnosis schema remains open-ended rather than limited to the example error causes.} The action includes: an error diagnosis and an actionable verbal instruction tailored to the inferred error cause (e.g., ``position yourself closer to the door before grasping the handle'').

\begin{figure}[!t]
  \centering
  \includegraphics[width=0.9\columnwidth]{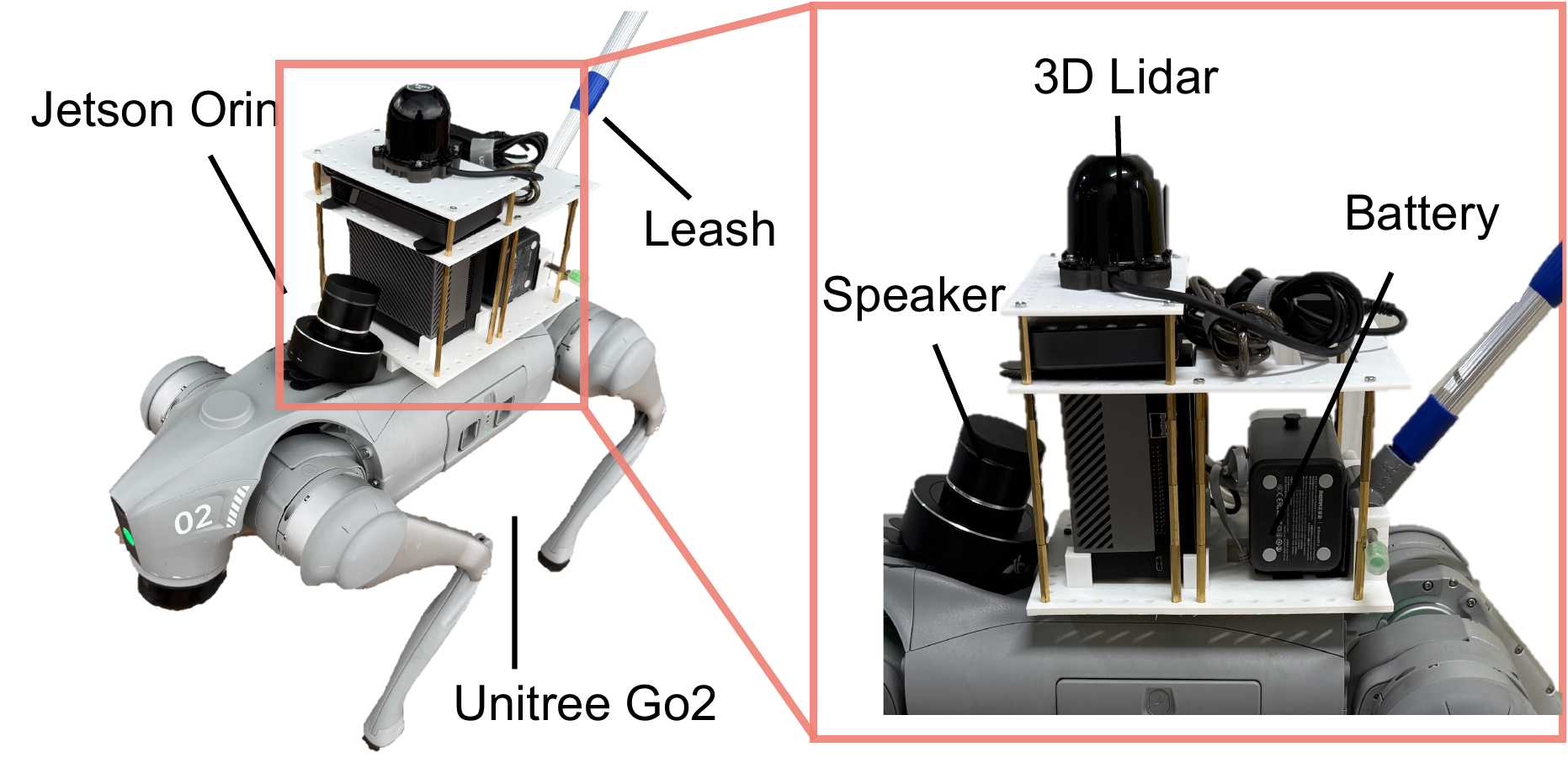}
  \caption{\textbf{Robot guide dog hardware setup.} Unitree Go2 quadruped robot equipped with Jetson AGX Orin for onboard computation, Hesai JT16 LiDAR for navigation, and chest-mounted camera on the user for first-person view analysis.}
  \label{fig:dog_setup}
    \vspace{-17pt}
\end{figure}

\noindent\textbf{Robot Adaptation ($f_{\text{param}}$).} This module takes as input the coaching action $c$ (which contains the error diagnosis and verbal instruction) and produces robot behavior adjustments $u$. Based on the diagnosed error cause, this module generates specific parameter updates (e.g., if the error is spatial misalignment due to positioning too far, adjust the robot's stopping position to a spot closer to the door or change the facing angle to make it easier for the human to grasp the handle). These parameter adjustments implement scaffolding: the robot adapts task difficulty to match the learner's proficiency, making the task easier when errors are detected and gradually increasing difficulty as proficiency improves.

\redtext{To reduce hallucinations, every stage is prompted to emit a constrained JSON object matching a predefined schema; outputs are parsed and validated before being passed downstream, with invalid or missing fields rejected and regenerated or replaced by conservative fallbacks. The frame stage restricts open-ended visual reasoning to explicit state fields, while later stages reason over structured summaries rather than raw free-form video descriptions.}

We instantiate $f_{\text{frame}}$ with a VLM that processes images, while $f_{\text{time}}$, $f_{\text{coach}}$, and $f_{\text{param}}$ use LLM with symbolic representations. \redtext{In practice, we use GPT-5.1 for all of them.} \bluetext{We use these models zero-shot with fixed prompts and JSON schemas, without fine-tuning on our study data.}

\noindent\textbf{Example.} Consider an episode where the user struggles to open the door efficiently. The VLM ($f_{\text{frame}}$) extracts frame-level states: ``t=0s: user approaching, door closed'', ``t=5s: user positioned far away from the door, hand searching'', ``t=12s: user grasped handle, door opening'', ``t=18s: door fully opened''. The timeline summarizer ($f_{\text{time}}$) aggregates this into: ``Duration: 18s, Outcome: Success, Timeline: t=0s to 5s, the user hesitated at the door for 3 seconds, t=5s to 12s, the user grasped the handle and opened the door, ...''. The coaching policy ($f_{\text{coach}}$) identifies the inefficiency cause (spatial misalignment: positioned too far) and generates: ``Position yourself closer to the door before grasping the handle—about arm's length away''. Finally, $f_{\text{param}}$ determines the appropriate robot adjustment: move stopping position closer to the door.

\tightsubsection{Hardware Platform}
The robot guide dog system (shown in~\figref{fig:dog_setup}) consists of a Unitree Go2 quadruped robot equipped with a Jetson AGX Orin for onboard computation and a Hesai JT16 LiDAR for environment sensing and obstacle detection. The robot provides autonomous navigation using a planner based on~\citet{xiao2021robotic} that guides users to specified goals while avoiding obstacles. Users interact with the robot through a rigid leash attachment. A DJI Action 6 camera mounted on the user's chest captures a first-person view for VLM-based behavior analysis, enabling the coaching system to observe the current state. \redtext{A microphone records the user feedback.} Generated instructions are delivered through a speaker.

\tightsection{Experiment}
\label{sec:rq}

We structure our experiment around the following research questions.
\begin{itemize}[leftmargin=*]
\item RQ1 (Component Feasibility): Are components of \name{} technically feasible for coaching? (\Secref{subsec:experiment_feasibility})
\item RQ2 (Human Subject Evaluation): Does \name{} achieve better objective learning outcomes and subjective user experience? (\Secref{subsec:human_subject})
\item RQ3 (Learning Retention): Do skills learned with \name{} persist over time? (\Secref{subsec:retention})
\item RQ4 (Case Study with a Visually Impaired User): What are the feasibility insights and additional design considerations with a visually impaired user? (\Secref{subsec:vi_case_study})
\end{itemize}
\begin{figure}[t!]
    \centering
    \includegraphics[width=\columnwidth]{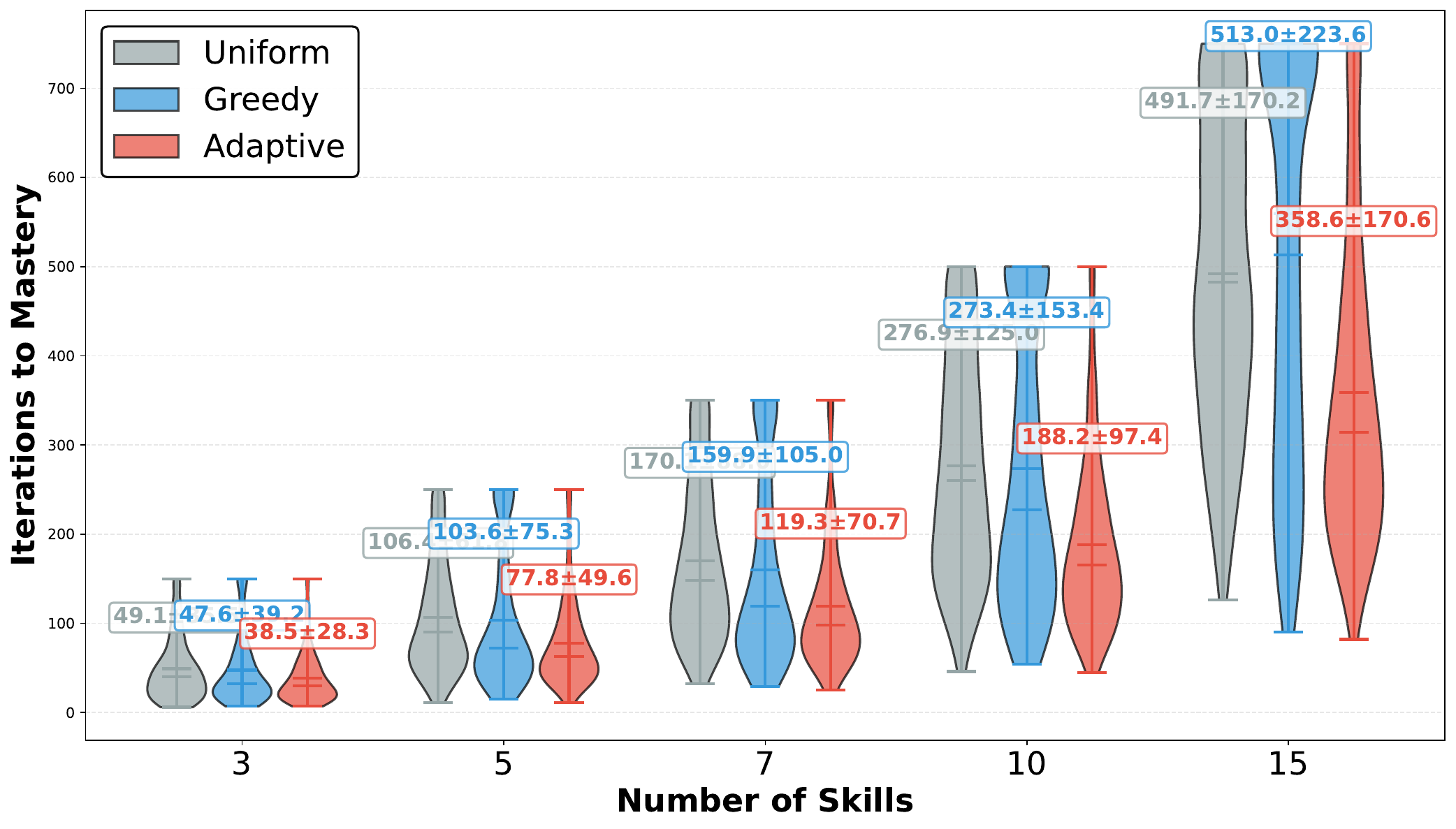}
    \caption{\textbf{Adaptive curriculum sequencing achieves faster learning.} Violin plots show the distribution of teaching actions needed to reach mastery for different numbers of sub-skills. Adaptive curriculum sequencing achieves faster learning.} 
    \label{fig:sim_user_curve}
    \vspace{-10pt}
\end{figure}

We address these questions across four studies. Study 1 validates the system's technical components, showing that our adaptive curriculum and decomposed reasoning pipeline outperform baselines in simulation and human judgment benchmarks. Study 2 demonstrates that \name{} significantly improves navigation performance and user perceived usefulness compared to general instruction in a controlled user study. Study 3 confirms that these learned skills persist after two weeks. Finally, Study 4 provides a case study with a visually impaired user, delivering design insights for deployment and understanding of extensibility to a handover task.

\tightsubsection{Study 1: Feasibility Evaluation}
\label{subsec:experiment_feasibility}  
To address RQ1, we evaluated the technical feasibility of \name{}'s key components.

\subsubsection{Simulated User Evaluation}
To evaluate adaptive curriculum sequencing, we developed a simulated learner following~\cite{Corbett2005KnowledgeTM, Wilson2016BackTT, pomdpteaching, lan2014sparse} capturing key learning dynamics (Power Law of Practice, slips/guesses, forgetting, fatigue; full details in Appendix~\ref{appendix:sim_human_model}). \bluetext{This allows us to compare curriculum policies across many learner trajectories before deploying them in the physical study.} We compared three curriculum strategies across 500 diverse learner profiles: \textbf{Uniform} (uniform sub-skill selection), \textbf{Greedy} (use the recent trial performance only to select sub-skill), and \textbf{Adaptive} (ours, maintains a cumulative proficiency model). As shown in~\figref{fig:sim_user_curve}, adaptive curriculum sequencing achieved faster learning compared to baselines.

\subsubsection{Frame Description Quality}
\begin{figure}[t]
    \centering
    \includegraphics[width=0.98\columnwidth]{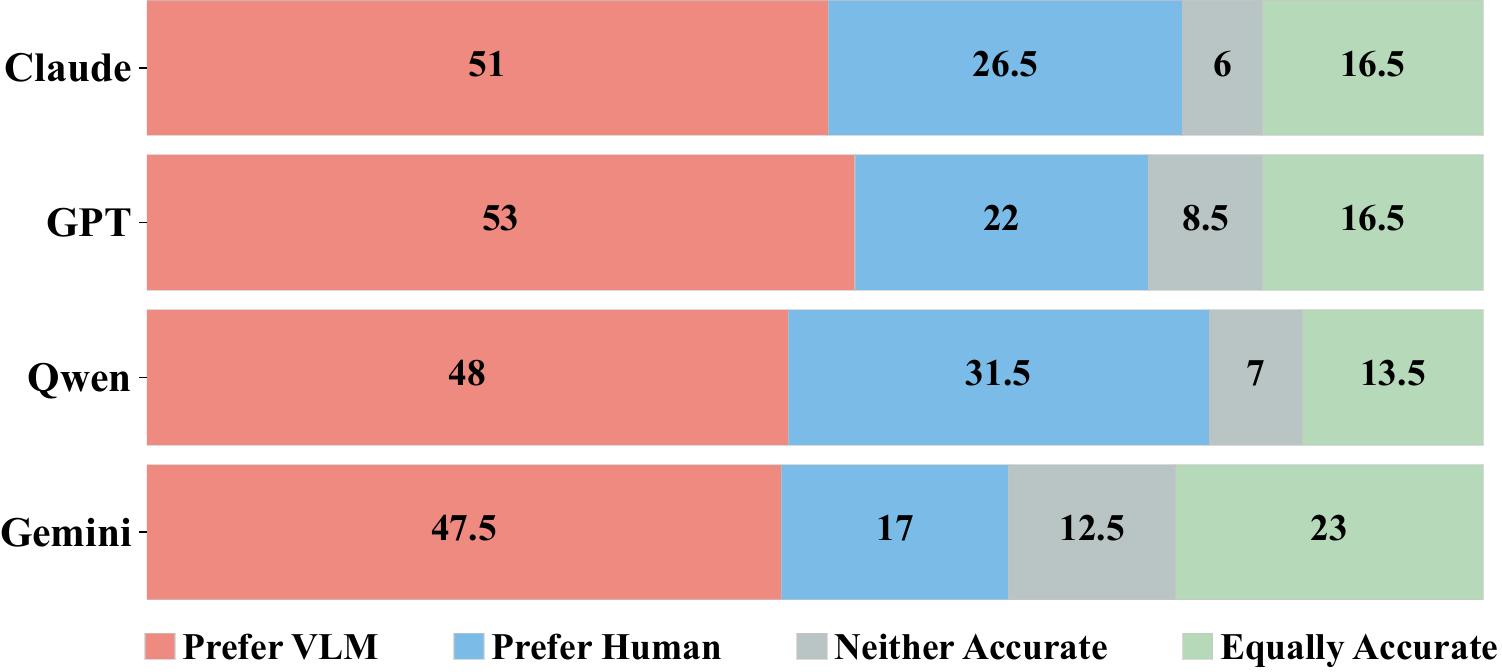}
    \caption{\textbf{Frame analysis accuracy comparison.} Preference rates of VLM-generated descriptions versus human annotations across different language models (GPT-5.1, Claude-4.5-Sonnet, Gemini-3.0-Pro amd Qwen3-VL).}
    \label{fig:human_judge}
    \vspace{-10pt}
\end{figure}

We evaluated whether VLMs can accurately extract high-quality state descriptions from video frames by comparing VLM predictions against human annotations on 100 frames. \redtext{Accurate frame descriptions are important because they are the evidence the coach uses to summarize an episode and diagnose causes of errors.} Independent human judges chose their preferred description through pairwise comparisons (\figref{fig:human_judge}) for the same frame. Given the evaluation results, we chose GPT-5.1 for our experiments.

\begin{table}[!t]
   \centering
   \setlength{\tabcolsep}{6pt}
   \caption{\textbf{Human evaluation on coaching feedback quality.} Preference rate indicates percentage of decisive preference. Other metrics are mean ratings on 1-5 Likert scales (higher is better). Latency is average time to generate feedback.
   }
   \begin{tabular}{lccc}
   \toprule
   Metric & \textbf{VLM} & \textbf{Structured VLM} & \textbf{Ours} \\
   \midrule
   Overall Preference Rate (\%) $\uparrow$ & 16.3 & 32.5 &  \cellcolor{lightblue}51.2 \\
   Helpfulness $\uparrow$ & 2.99 & 3.49 & \cellcolor{lightblue}3.66 \\
   Factuality $\uparrow$ & 3.26 & 3.50 & \cellcolor{lightblue}3.59 \\
   Specificity $\uparrow$ & 3.18 & \cellcolor{lightblue}3.95 & 3.91 \\
   \midrule
   Latency (s) $\downarrow$ & 44.36 & 42.15 & \cellcolor{lightblue}20.62 \\
   \bottomrule
   \end{tabular}
   \label{tab:simple_vlm}
   \vspace{-7pt}
\end{table}

\subsubsection{Comparison of Different VLM Pipelines}
While directly processing videos with VLMs could simplify the architecture, it leads to prolonged waiting times that disrupt interactive coaching and fails on longer episodes due to context length constraints. To validate our decomposed approach, we compared it against two alternatives on 10 navigation videos: (1) \textbf{VLM}: direct video prompting, (2) \textbf{Structured VLM}: full video input with structured prompt similar to \name{}, and (3) \textbf{Ours}: frame-level decomposition. \redtext{This comparison evaluates whether the staged representation used by \name{} preserves enough episode context to generate useful coaching while reducing latency.} Eight human judges evaluated feedback quality through overall preference rankings, with additional Likert-scale ratings on three supporting dimensions: \textit{Helpfulness} (overall utility for learning), \textit{Factuality} (accuracy of observations), and \textit{Specificity} (level of actionable detail) (details in Appendix~\ref{appendix:vlm_comparison}). \redtext{We report four metrics in \tabref{tab:simple_vlm}: Overall Preference Rate is the percentage of decisive judge preferences won by each method; Helpfulness, Factuality, and Specificity are mean 1--5 Likert ratings, where higher is better; and Latency is the average wall-clock time in seconds from video input to generated feedback on the same hardware and API endpoints.} Instructions generated by \name{} were preferred in 51.2\% of comparisons while achieving faster generation as shown in~\tabref{tab:simple_vlm}. Having validated the technical components, we next evaluate the complete system with human participants.
\redtext{The mean end-to-end latency of our decomposed pipeline was 20.62\,s, which supports terminal feedback between practice episodes; the full latency breakdown is provided in the supplementary material.}

\tightsubsection{Study 2: Human Subject Evaluation}
\label{subsec:human_subject}
To address RQ2, we evaluated \name{} through a two-stage approach: \redtext{a controlled between-subject study with blindfolded sighted participants as a proxy} followed by an exploratory case study with a visually impaired participant (\Secref{subsec:vi_case_study}).
\begin{figure}[!t]
  \centering
  \includegraphics[width=0.98\columnwidth]{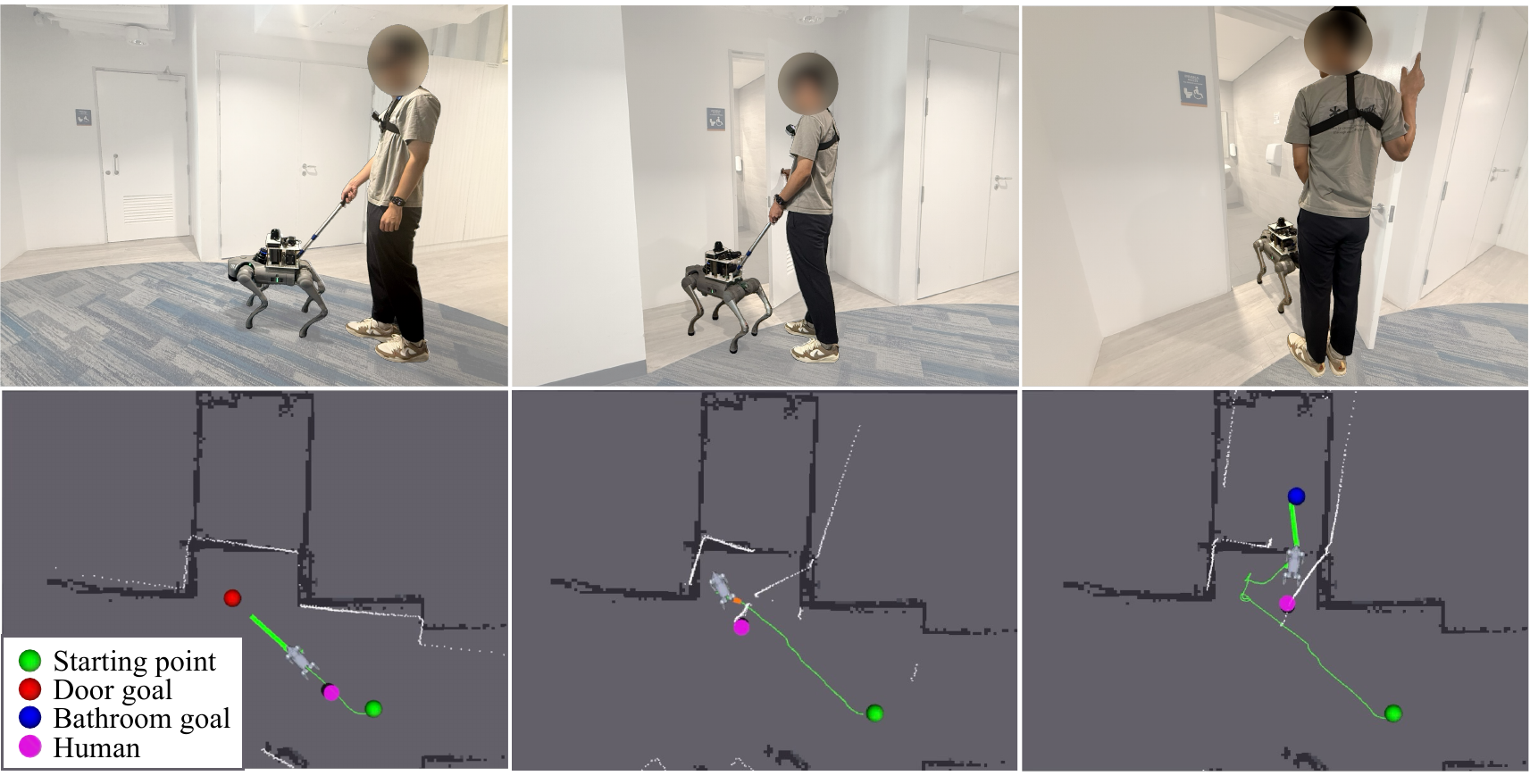}
  \vspace{-2pt}
  \caption{\textbf{Navigation task setup and experimental environment.} We show the three sub-skills of door navigation: Navigate to Door, Open Door, and Enter Room. \textbf{Top}: human behavior. \textbf{Bottom}: visualization of the navigation map.}
  \label{fig:dog_setup_vis}
  \vspace{-15pt}
\end{figure}

\noindent\textbf{Task Description.} Participants navigated a doorway using the robot guide dog (\figref{fig:dog_setup_vis}). The task requires mastering the three sequential sub-skills defined in \Secref{sec:method}: (1) Navigate to Door, (2) Open Door, and (3) Enter Room.

\noindent\textbf{Study Design.} We recruited 22 blindfolded sighted participants ($M=27, SD=3.66$, 5 females): 2 for pilot, 20 for the main between-subjects study comparing \name{} against a baseline providing task-level verbal instructions without sub-skill decomposition, personalized feedback, or adaptive sequencing (full details in Appendix~\ref{appendix:baseline_instruction}). \redtext{We use blindfolded participants to obtain controlled quantitative evidence and interpret this population as a proxy rather than as a substitute for statistically powered evaluation with VI users.}

\subsubsection{Pilot Study}
We conducted a pilot study with 2 participants to establish optimal performance, identify common failure modes, and refine the experimental protocol. 

\subsubsection{Main Study}
Following the pilot study, we recruited 20 blindfolded sighted participants for the main study, randomly assigned to one of the two teaching methods (10 participants per group). All participants gave written informed consent and were compensated for their time. None of the participants had prior experience using the robot guide dog. The study was IRB-approved.

\noindent\textbf{Experimental Protocol.} Each participant completed a single session lasting approximately 40 minutes. The session consisted of five phases: Introduction (5 min), Practice (2 trials, 7 min), Training (11 min), Evaluation (2 trials, 7 min), and Survey (10 min). During the Training phase, the baseline group performed 3 full end-to-end trials. The \name{} group practiced up to 6 sub-skill training episodes in total selected by the adaptive curriculum, with the session ending early if the system determined all sub-skills were mastered. 

\noindent\textbf{Evaluation Metrics.} Objective: Time to complete route (in seconds), averaged across the two evaluation trials. Subjective: NASA Task Load Index (NASA-TLX)~\cite{Hart1988DevelopmentON}, System Usability Scale (SUS)~\cite{Brooke1996SUSA}, Customized Questionnaires for Perceived Usefulness (PU) and Perceived Ease of Use (PEU) (full questionnaire in Appendix~\ref{appendix:usefulness_questionnaire}). 
\begin{table}[!t]
\centering
\vspace{-8pt}
\renewcommand{\arraystretch}{0.3}
\setlength{\tabcolsep}{1pt}
\begin{tabular}{cc}
\hspace{-10pt}\includegraphics[width=0.53\columnwidth]{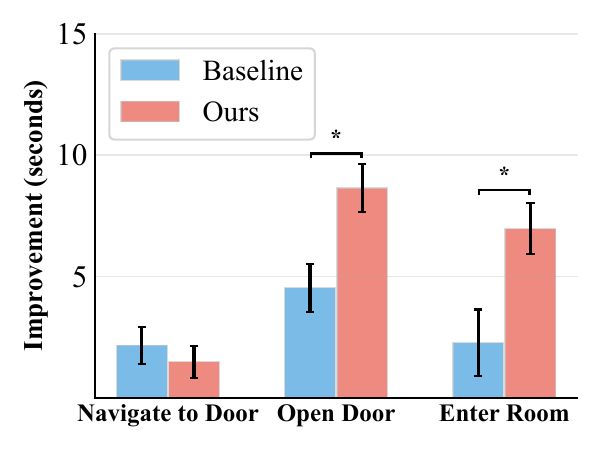} & \hspace{-7pt}
\includegraphics[width=0.53\columnwidth]{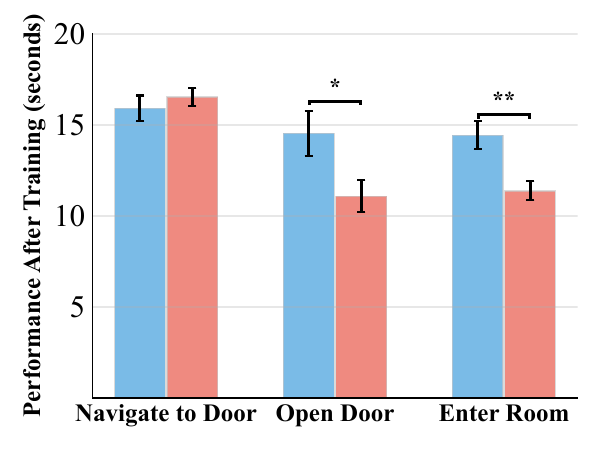} \\
\textit{a}) & \textit{b})
\end{tabular}
\vspace{-5pt}
\captionof{figure}{\textbf{Objective performance.} \textit{a}) Performance improvement across sub-skills. \textit{b}) Final completion time in evaluation trials.}
\label{fig:dog_objective}
\vspace{-10pt}
\end{table}

\begin{figure}[!t]
    \centering
    \includegraphics[width=\columnwidth]{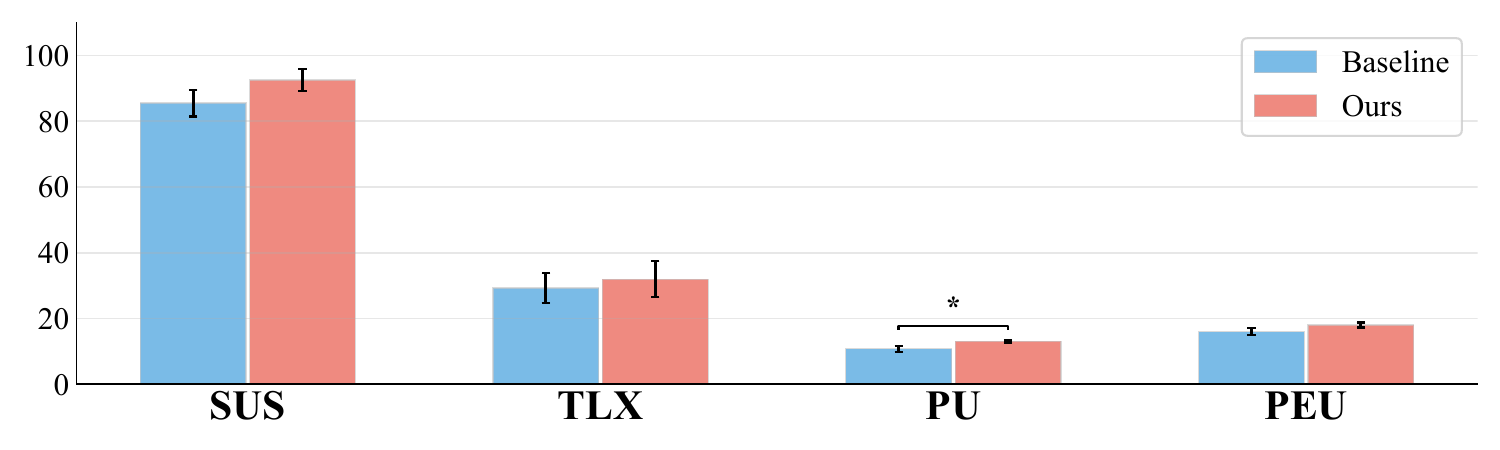}
    \vspace{-20pt}
    \caption{\textbf{Subjective evaluation results.} Participants rated \name{} as having significantly higher Perceived Usefulness (PU) compared to general instruction.}
    \label{fig:dog_subjective}
    \vspace{-20pt}
\end{figure}

\begin{figure*}[!t]
\centering
\includegraphics[width=0.98\textwidth]{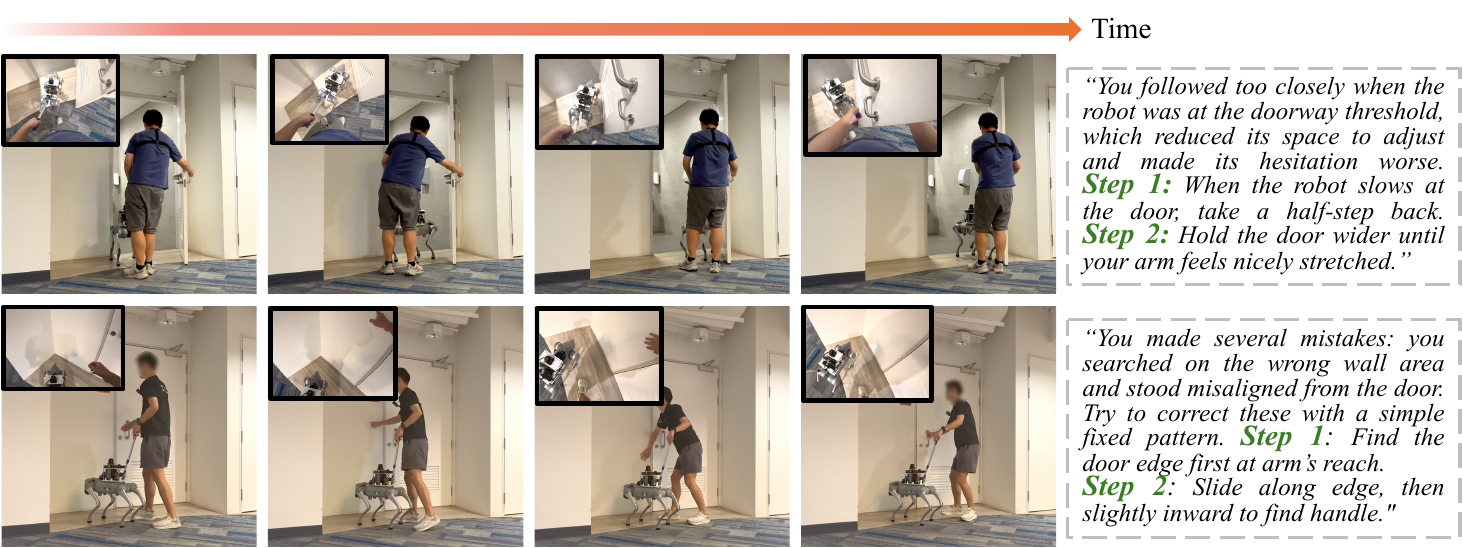}
\caption{\textbf{Qualitative examples of generated instructions.} Top-left shows the image captured by the chest-mount camera.}
\vspace{-13pt}
\label{fig:qualitative_navigation}
\end{figure*}

\begin{figure}[!t]
  \centering
  \includegraphics[width=\columnwidth]{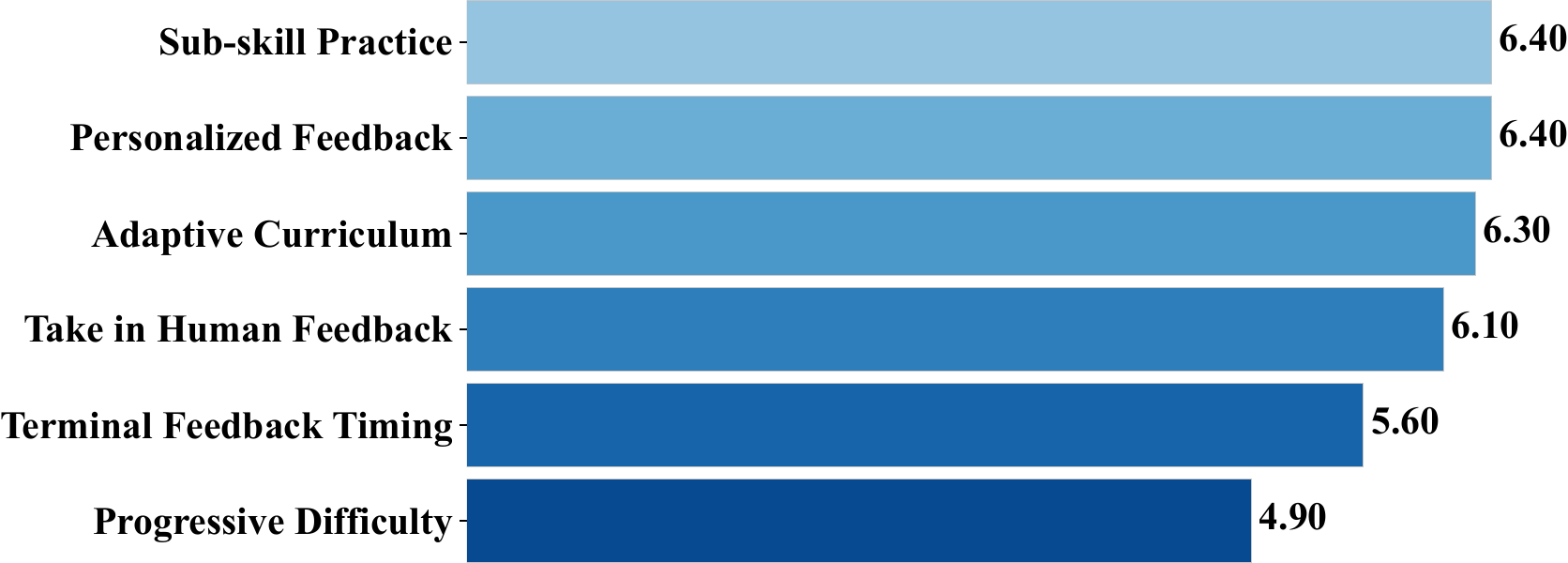}
  \caption{\textbf{User ratings of \name{} components.} Using a 1 - 7 Likert scale (1 = not helpful, 7 = very helpful).}
  \vspace{-7pt}
  \label{fig:dog_preference}
\end{figure}

\subsubsection{Results on Objective Performance}
\emph{Coaching significantly improves learning rates and final performance for interaction-intensive sub-skills}
As shown in \figref{fig:dog_objective} (all comparisons use Welch's t-test; all times in seconds), for the ``Navigate to Door'' sub-skill, no significant difference was observed. This is primarily because the task involves straightforward following of the robot's guidance without requiring complex coordination or spatial reasoning. Thus, we focus our analysis on the interactive parts of the task. Coaching is particularly effective for these coordination tasks. For the ``Open Door'' task, the experimental group showed significantly greater improvement from pre-training to post-training (improvement: $M = 8.64, SD = 0.99$ vs. baseline $M = 4.53, SD = 1.00$), $t(18) = -2.78, p = .012, g = -1.19$, resulting in faster final completion times ($M = 11.07, SD = 0.89$) compared to baseline ($M = 14.54, SD = 1.23$), $t(18) = 2.12, p = .045, g = 0.93$. Similarly, for ``Enter Room,'' the \name{} group achieved greater improvement ($M = 6.96, SD = 1.05$ vs. $M = 2.27, SD = 1.37$), $t(18) = -2.61, p = .019, g = -1.11$, leading to faster completion times ($M = 11.36, SD = 0.52$ vs. $M = 14.42, SD = 0.76$), $t(18) = 3.00, p = .006, g = 1.35$.
\subsubsection{Results on Subjective Experience}
\emph{Participants rated the coaching system as more useful than baseline instruction, valuing its targeted feedback and adaptive structure.}
As shown in~\figref{fig:dog_subjective}, participants rated \name{} as having significantly higher Perceived Usefulness ($M = 13.10, SD = 1.29$ vs. $M = 10.80, SD = 2.82$), $p = .036, g = 1.05$. General usability metrics (SUS, Workload) showed no significant differences; we hypothesize that these general measures primarily capture participants' experience with the overall robot guide dog system rather than the coaching system specifically.

Thus, we examined which coaching features participants found most helpful (\figref{fig:dog_preference}). \redtext{After training, participants rated each \name{} component on a 1--7 Likert scale, where 1 indicated ``not helpful'' and 7 indicated ``very helpful.''} The most appreciated features were: (1) dedicated sub-skill training, (2) personalized feedback, and (3) adaptive curriculum.  They reported that they learned skills (e.g., ``Try to align with the robot dog direction during movement.'') from personalized feedback. \redtext{This pattern is consistent with the component studies: the simulated-learner results suggest that adapting practice to estimated proficiency improves curriculum selection, while the pipeline comparison suggests that the decomposed feedback generator can produce useful episode-level coaching. In the full system, participants most clearly noticed the human-facing parts of this design: targeted sub-skill practice, personalized feedback, and adaptive sequencing. Robot-side adaptation acted mainly as conservative scaffolding through task-difficulty adjustments, which may explain why blindfolded participants found it less salient than the feedback and curriculum components.} 

Following the main experiment, we conducted a post-hoc comparison with the same 10 participants from the \name{} group, where a human expert coached them on the same task. When asked for their preference, 70\% of participants favored the human expert, citing two key advantages: (1) the ability to provide physical corrections (e.g., manually adjusting body posture), and (2) the delivery of concurrent, real-time feedback that addresses questions and offers instructive guidance during execution. \bluetext{This preference also clarifies the role of our terminal-feedback design: it supports reflection between attempts, but does not correct body pose or timing while the user is acting.} Conversely, two participants preferred the robot coach, valuing its consistent, standardized feedback and the sense of autonomy and privacy it offered. One participant remained neutral. Qualitative examples can be found in~\figref{fig:qualitative_navigation}.

\tightsubsection{Study 3: Retention Analysis}
\label{subsec:retention}
To address RQ3, we conducted follow-up sessions with a subset of participants (N=5) from the \name{} group in Study 2, two weeks after their training. This within-subject analysis provides preliminary evidence of whether skills learned with \name{} persist over time. \redtext{Because only five participants returned, we interpret the retention study as preliminary longitudinal evidence rather than a definitive estimate of long-term retention.} Participants completed both the same routes from the evaluation phase of Study 2 and novel routes with different layouts to assess temporal retention.
\subsubsection{Results}
\emph{Skills learned with coaching persist over two weeks, with complex coordination skills showing continued improvement and simpler skills remaining stable.}

\begin{figure}[!t]
    \centering
    \includegraphics[width=\columnwidth]{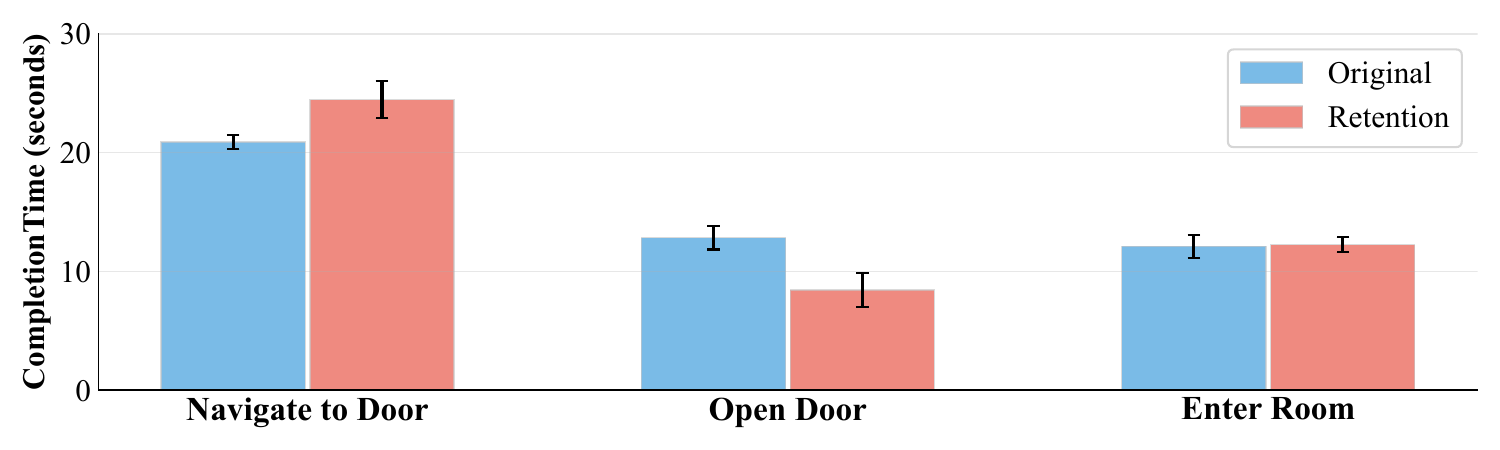}
    \vspace{-20pt}
    \caption{\textbf{Skill retention.} Bars show the mean retention change for each skill in second.}
    \label{fig:dog_retention}
    \vspace{-15pt}
\end{figure}

As illustrated in~\figref{fig:dog_retention}, performance at the two-week retention session varied by sub-skill. Participants demonstrated substantial improvement in the ``Open Door'' task. In contrast, ``Enter Room'' and ``Navigate to Door'' showed minor performance degradation, indicating these skills remained relatively stable. Interestingly, 60\% of participants identified ``Open Door'' as the most challenging aspect of the task. We hypothesize that the observed improvement in this difficult sub-skill reflects the incubation effect~\cite{Sio2009DoesIE}, where complex motor skills continue to consolidate during rest periods as the brain integrates and refines learned patterns.

\tightsubsection{Study 4: Case Study with Visually Impaired User}
\label{subsec:vi_case_study}
To address RQ4, we conducted an exploratory case study with a totally blind participant (Male, 70, with no residual vision) to identify design considerations and evaluate system usability for navigation and extensibility to robotic handover tasks (details in Appendix~\ref{appendix:vi_protocol} and Appendix~\ref{appendix:handover_setup}).

\subsubsection{Convergence with Study 2}
Training with \name{} yielded clear learning progression, with the participant's positive feedback (\textit{``The feedback is very accurate.''}). \redtext{The participant completed the same navigation coaching-session structure as the controlled-study participants and improved ``Open Door'' completion time from 22.4\,s to 9.1\,s, a trend consistent with Study 2.} \bluetext{This session followed the same structure as Study 2, with two initial practice trials, up to six coached sub-skill episodes, and two evaluation trials.} Preferences aligned remarkably with Study 2: the participant independently ranked sub-skill practice, personalized feedback, and adaptive curriculum as the top features, and also confirmed ``Open Door'' as the most challenging sub-skill.

\subsubsection{VI-Specific Designs}
However, the study revealed critical distinctions. Unlike blindfolded proxies who may rely on residual visual-spatial mental models, the VI participant emphasized a need for \textit{multimodal grounding}, specifically haptic cues to supplement verbal feedback and provide spatial information that language alone cannot convey. In addition, the VI user requested greater control over the coaching system, specifically the ability to pause, replay, and skip feedback to manage cognitive load. 

\subsubsection{Extensibility to Robotic Handover}
\begin{figure}[t]
  \centering
  \includegraphics[width=0.8\columnwidth]{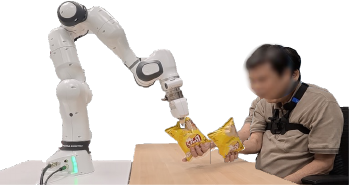}
  \caption{\textbf{Experimental setup for the handover task.} Human practices receiving objects from a robot manipulator.}
  \label{fig:handover_setup}
  \vspace{-16pt}
\end{figure}

We applied \name{} to a robotic handover task~\citep{strabala2013toward, chan2013human, kupcsik2017learning} as shown in~\figref{fig:handover_setup}. In this task, a robotic arm hands an object to a human, requiring human hand positioning and grasp timing without vision. \bluetext{We use this study as an instantiation showing that the coaching pipeline can be configured for another interaction task.} Cross-task comparison revealed task-dependent coaching needs: while terminal feedback succeeded for navigation, the participant required \textit{concurrent} (real-time) feedback for handover to adjust hand position during execution. This suggests coaching systems must adapt not only feedback content but also its timing based on task dynamics.
\redtext{We treat this handover result as an extensibility demonstration rather than a full evaluation: it shows that the architecture can be instantiated for a different interactive assistance task, but does not establish task-general effectiveness without a larger controlled study.}

\section{Conclusion}

We presented \name{}, an automated coaching system designed to teach visually impaired users how to navigate with a robot guide dog. \name{} dynamically sequences sub-skills and generates personalized, context-aware feedback. By leveraging foundation models and knowledge tracing, our system delivers a dynamic curriculum and actionable guidance. Evaluations with simulated agents and human participants demonstrate that \name{} accelerates human learning in complex coordination tasks compared to standard instruction, with skill retention persisting over two weeks.

Our current approach relies on expert-defined reasoning schemas to ensure safety and interpretability, a necessary trade-off that limits fully open-ended adaptation. Furthermore, while coaching is automated, the system relies on expert-defined skill decomposition, and physical safety and task resetting still require human oversight. \redtext{The current study has four limitations: terminal feedback avoids cognitive overload during safety-critical navigation but cannot support in-the-moment correction; the chest-mounted camera can suffer from occlusion, blur, or viewpoint limitations; robot-side adaptation was evaluated as conservative scaffolding rather than through a full factorial ablation; and the visually impaired case study plus five-participant retention study provide preliminary evidence rather than population-level validation.} \bluetext{Longer deployments are needed to study mutual adaptation over time.} Future work will focus on longitudinal deployment with the visually impaired community to validate real-world efficacy, developing adaptive feedback mechanisms that transition from terminal to concurrent guidance based on task demands, \redtext{using multi-camera sensing or explicit body-orientation guidance,} automating skill discovery to reduce reliance on expert-defined sub-skill decomposition, \redtext{conducting ablations of system components,} \bluetext{including human-selected curriculum baselines,} and extending \name{} to more complex interactive navigation tasks such as elevator use, escalators, and navigation in crowded environments.

\section*{Acknowledgments}

This research is supported in part by the National Research Foundation, Singapore under its AI Singapore
Programme under AISG Award AISG2-PhD-2022-01-036[T] by Ministry of Education Academic Research Fund (grant no. T1-251RES2406), and by Ministry of Digital Information under its AIVP program. We thank the anonymous reviewers for their valuable feedback, and our participants for their time and insights. We also thank Shaojun Cai for his support and discussions that helped shape this work.

\bibliographystyle{plainnat}
\bibliography{ref}

\onecolumn

\appendix

\ifdefined\standaloneappendix
\tableofcontents
\vspace{1em}
\fi

\newcommand{\timestep}{t}     %
\newcommand{\modality}{m}     %

\newcommand{\performance}{y}  %
\newcommand{\opttime}{\tau^{*}}
\newcommand{\basemean}{\hat{\mu}}
\newcommand{\basesigma}{\hat{\sigma}}
\newcommand{\currentmean}{\mu}
\newcommand{\currentsigma}{\sigma}

\newcommand{\learnprogress}{p}  %
\newcommand{\attemptcount}{N}   %
\newcommand{\assistflag}{a}     %

\newcommand{\learnrate}{\alpha}
\newcommand{\forgetrate}{\lambda_{\text{forget}}}
\newcommand{\fatiguethresh}{N_{\text{fatigue}}}
\newcommand{\fatiguerate}{\lambda_{\text{fatigue}}}
\newcommand{\experiencerate}{\lambda_{\text{exp}}}

\newcommand{\assisteffect}{\eta}
\newcommand{\diminishfactor}{\gamma_{\text{dim}}}

\subsection{Study 1: Simulated Learner Model}
\label{appendix:sim_human_model}
This section provides the formal mathematical specification of the simulated learner model used to evaluate the adaptive teaching policy. The model integrates key cognitive factors grounded in established learning theories and prior work on simulated learners for intelligent tutoring systems~\cite{Corbett2005KnowledgeTM, Wilson2016BackTT, pomdpteaching, lan2014sparse}.

\noindent\textbf{Model Overview.} The simulated learner maintains a latent proficiency state for each sub-skill that evolves over time through practice. At each teaching step $\timestep$, when the learner practices a sub-skill $\skill$, the model: (1) computes the current expected performance time $\currentmean_{\skill,\timestep}$ based on accumulated learning progress, forgetting, and fatigue; (2) samples an observed performance time $\performance_{\skill,\timestep}$ from a Gaussian distribution to simulate behavioral variability; (3) updates the learning progress $\learnprogress_{\skill,\timestep}$ to reflect skill improvement from practice. This creates realistic learning trajectories that exhibit improvement with practice, performance variability, skill decay without practice, and fatigue from over-practice.

\noindent\textbf{Overview of Cognitive Factors} The simulated learner reflects the multifaceted and stochastic nature of human learning through several key factors:
\begin{itemize}
\item \textbf{Power Law of Practice:} Learners improve proficiency through practice following the Power Law of Practice, where performance time decreases as a power function of the number of practice trials. This captures the well-established phenomenon that early practice yields rapid improvement, with diminishing gains as proficiency approaches mastery.
\item \textbf{Slips and Guesses:} Following Bayesian Knowledge Tracing, the model accounts for slips (errors on mastered skills due to momentary lapses) and guesses (correct responses on unmastered skills due to chance), reflecting stochastic variability in behavior. This is implemented through Gaussian noise in performance observations.
\item \textbf{Forgetting:} Performance degrades over time without practice. The forgetting rate $\lambda_{\text{forget}}$ determines how quickly skills decay when not practiced.
\item \textbf{Fatigue Effects:} Performance degradation from extended practice sessions is modeled following cognitive load theory. After a threshold number of attempts $N_{\text{fatigue}}$, additional practice within the same session yields diminishing returns and may temporarily impair performance.
\item \textbf{Individual Variability:} Learners exhibit heterogeneous learning rates, initial proficiency levels, and susceptibility to fatigue. We model this by sampling learner parameters from distributions (Table~\ref{tab:sim_params}) to create diverse learner profiles.
\end{itemize}
\noindent\textbf{Model Initialization}
For each skill $\skill$, the learner's initial state is defined by a baseline mean performance ($\basemean_{\skill}$) and standard deviation ($\basesigma_{\skill}$). These are derived from the optimal performance time $\opttime_{\skill}$, a performance multiplier $m_{\skill} > 1$, and a variance factor $v_{\skill} > 0$.
\begin{align}
\basemean_{\skill} &= \opttime_{\skill} \cdot m_{\skill} \\
\basesigma_{\skill} &= \opttime_{\skill} \cdot v_{\skill}
\end{align}
Table~\ref{tab:sim_params} provides the distributions used to initialize these parameters for diverse learner profiles.
\begin{table}[t]
\centering
\caption{Parameter distributions for the Simulated Learner. Parameters are randomized per skill or per learner episode to create diverse student profiles.}
\label{tab:sim_params}
\small
\begin{tabular}{p{4cm} p{5cm} l}
\toprule
\textbf{Parameter} & \textbf{Distribution / Logic} & \textbf{Description} \\
\midrule
\multirow{3}{*}{Optimal Performance Time ($\opttime_{\skill}$)} & 30\%: $\mathcal{U}(1.0, 5.0)$ s (Easy) & \multirow{3}{*}{Minimum time to complete skill} \\
 & 30\%: $\mathcal{U}(5.0, 12.0)$ s (Medium) & \\
 & 40\%: $\mathcal{U}(12.0, 25.0)$ s (Hard) & \\
\midrule
\multirow{3}{*}{Initial Multiplier ($m_{\skill}$)} & 20\%: $\mathcal{U}(1.0, 2.0)$ (Near Expert) & \multirow{3}{*}{Initial incompetence factor ($T_{init} = m \cdot T_{opt}$)} \\
 & 30\%: $\mathcal{U}(2.0, 5.0)$ (Novice) & \\
 & 50\%: $\mathcal{U}(5.0, 15.0)$ (Struggling) & \\
\midrule
\multirow{2}{*}{Variance Factor ($v_{\skill}$)} & 30\%: $\mathcal{U}(0.05, 0.2)$ (Consistent) & \multirow{2}{*}{Performance noise magnitude} \\
 & 70\%: $\mathcal{U}(0.3, 1.2)$ (Noisy) & \\
\midrule
\multirow{3}{*}{Learning Rate ($\alpha_{\skill}$)} & 25\%: $\mathcal{U}(0.2, 0.5)$ (Fast) & \multirow{3}{*}{Rate of proficiency gain per attempt} \\
 & 25\%: $\mathcal{U}(0.01, 0.08)$ (Slow) & \\
 & 50\%: $\mathcal{U}(0.08, 0.25)$ (Average) & \\
\midrule
Forgetting Rate ($\phi$) & $\mathcal{U}(0.02, 0.1)$ & Proficiency decay per non-practice step \\
\midrule
Fatigue Threshold & $\mathcal{U}_{int}(2, 20)$ attempts & Steps before performance degrades \\
\bottomrule
\end{tabular}
\end{table}

Initially, at $\timestep=0$, the learning progress $\learnprogress_{\skill,0}$ and attempt count $\attemptcount_{\skill,0}$ for all skills are zero.

\noindent\textbf{Performance on an Attempt}
At any given time step $\timestep$, the learner's performance $\performance_{\skill,\timestep}$ on a task for skill $\skill$ is modeled as a random variable drawn from a Gaussian distribution. The distribution is defined by a time-varying mean $\currentmean_{\skill,\timestep}$ (representing the learner's true underlying skill level) and standard deviation $\currentsigma_{\skill,\timestep}$ (representing performance variability due to slips and guesses).
\begin{equation}
\performance_{\skill,\timestep} \sim \mathcal{N}(\currentmean_{\skill,\timestep}, \currentsigma_{\skill,\timestep}^2)
\end{equation}
The final performance time is clipped to a minimum positive value to ensure physical plausibility, i.e., $\max(0.1, \performance_{\skill,\timestep})$.
The current mean performance, $\currentmean_{\skill,\timestep}$, represents the learner's expected proficiency. It dynamically updates based on learning progress, forgetting, and fatigue. The equation starts from the baseline novice performance $\basemean_{\skill}$, subtracts improvements from learning ($I_{\text{learn}}$) and recovery from forgetting ($I_{\text{forget}}$), and adds penalties from fatigue ($I_{\text{fatigue}}$). The result is clipped to never go below the optimal expert time $\opttime_{\skill}$.
\begin{equation}
\currentmean_{\skill,\timestep} = \max\left(\opttime_{\skill}, \basemean_{\skill} - I_{\text{learn}} - I_{\text{forget}} + I_{\text{fatigue}}\right)
\end{equation}
where:
\begin{itemize}
\item \textbf{Learning Improvement ($I_{\text{learn}}$)}: The reduction in performance time due to accumulated knowledge. The learning progress $\learnprogress_{\skill,\timestep-1}$ ranges from 0 (no learning) to 1 (full mastery) and is updated after each practice attempt (see Learning Update Mechanism below).
\begin{equation}
I_{\text{learn}} = \learnprogress_{\skill,\timestep-1} \cdot (\basemean_{\skill} - \opttime_{\skill})
\end{equation}
This establishes a linear relationship between learning progress and expected performance: when $\learnprogress_{\skill,\timestep-1} = 0$, the learner performs at baseline novice level ($\currentmean_{\skill,\timestep} = \basemean_{\skill}$), and when $\learnprogress_{\skill,\timestep-1} = 1$, the learner achieves optimal expert performance ($\currentmean_{\skill,\timestep} = \opttime_{\skill}$).
\item \textbf{Forgetting Penalty ($I_{\text{forget}}$)}: Performance degradation due to elapsed time $\Delta \timestep$ since the last attempt.
\begin{equation}
I_{\text{forget}} = \forgetrate \cdot \Delta \timestep \cdot (\basemean_{\skill} - \opttime_{\skill})
\end{equation}
\item \textbf{Fatigue Penalty ($I_{\text{fatigue}}$)}: Performance degradation from excessive recent practice. This penalty is applied only when the attempt count exceeds a threshold $\fatiguethresh$.
\begin{equation}
I_{\text{fatigue}} = \fatiguerate \cdot \max(0, \attemptcount_{\skill,\timestep-1} - \fatiguethresh) \cdot (\basemean_{\skill} - \opttime_{\skill})
\end{equation}
\end{itemize}

\noindent\textbf{Current Performance Variance}
The variance $\currentsigma_{\skill,\timestep}$ models the consistency of the learner's performance. It decreases as the learner gains more experience, reflecting more stable performance.
\begin{equation}
\currentsigma_{\skill,\timestep} = \max\left(0.1 \cdot \opttime_{\skill}, \basesigma_{\skill} \cdot (1 - f_{\text{exp}})\right)
\end{equation}
where the experience factor $f_{\text{exp}}$ is given by:
\begin{equation}
f_{\text{exp}} = \min(0.5, \experiencerate \cdot \attemptcount_{\skill,\timestep-1})
\end{equation}
\noindent\textbf{Learning Update Mechanism}
After each attempt, the learner's progress $\learnprogress_{\skill,\timestep}$ is updated. This updated value is then used in the next attempt to compute $I_{\text{learn}}$ (see equation above), which determines how much the learner's expected performance has improved. The model assumes learning always occurs with practice, with the magnitude of the update implementing the Power Law of Practice through diminishing returns.
\begin{equation}
\learnprogress_{\skill,\timestep} = \min\left(1, \learnprogress_{\skill,\timestep-1} + \Delta\learnprogress_{\skill,\timestep}\right)
\end{equation}
The learning increment, $\Delta\learnprogress_{\skill,\timestep}$, is calculated as:
\begin{equation}
\Delta\learnprogress_{\skill,\timestep} = \max\left(0.01, \learnrate_{\skill} \cdot (1 - \diminishfactor \learnprogress_{\skill,\timestep-1})\right)
\end{equation}
where:
\begin{itemize}
\item $\learnrate_{\skill}$ is the base learning rate for skill $\skill$, sampled from a distribution to model individual differences.
\item $\diminishfactor$ is a coefficient modeling diminishing returns as mastery increases.
\item The minimum value of 0.01 ensures some learning always occurs, preventing complete stagnation.
\end{itemize}
This formulation captures the empirically observed pattern that novices improve rapidly with initial practice, while experts require substantially more practice to achieve marginal improvements~\cite{Wilson2016BackTT}.

\subsection{Study 1: Simulated User Evaluation Results}
\label{appendix:sim_user_results}

This section presents detailed results from the simulated learner evaluation comparing three curriculum sequencing strategies: Uniform, Greedy, and Adaptive. We evaluate each strategy's ability to help diverse learners master all sub-skills within a fixed budget of 50 teaching actions per skill.

\begin{table}[!t]
    \centering
    \setlength{\tabcolsep}{8pt}
    \caption{Curriculum sequencing performance on simulated learners. Mean±Std shows average teaching actions used, and Success rate (\%) for mastering all sub-skills within 200 teaching actions across varying curriculum complexity.}
    \begin{tabular}{ccccccc}
        \toprule
        \multirow{2}{*}{\textbf{\#Sub-skills}} & \multicolumn{2}{c}{\textbf{Uniform}} & \multicolumn{2}{c}{\textbf{Greedy}} & \multicolumn{2}{c}{\textbf{Adaptive}} \\
        \cmidrule(lr){2-3} \cmidrule(lr){4-5} \cmidrule(lr){6-7}
        & \textbf{Mean±Std} & \textbf{Success} & \textbf{Mean±Std} & \textbf{Success} & \textbf{Mean±Std} & \textbf{Success} \\
        \hline
        3 & 49.1±35.5 & 97.6\% & 47.6±39.2 & 93.8\% & \cellcolor{lightblue}38.5±28.3 & \cellcolor{lightblue}98.6\% \\
        5 & 106.4±61.8 & 96.8\% & 103.6±75.3 & 87.4\% & \cellcolor{lightblue}77.8±49.6 & \cellcolor{lightblue}98.0\% \\
        7 & 170.1±88.0 & 93.6\% & 159.9±105.0 & 86.4\% & \cellcolor{lightblue}119.3±70.7 & \cellcolor{lightblue}97.0\% \\
        10 & 276.9±125.0 & 93.0\% & 273.4±153.4 & 78.4\% & \cellcolor{lightblue}188.2±97.4 & \cellcolor{lightblue}98.0\% \\
        15 & 491.7±170.2 & 90.2\% & 513.0±223.6 & 66.0\% & \cellcolor{lightblue}358.6±170.6 & \cellcolor{lightblue}93.6\% \\
        \bottomrule
    \end{tabular}
    \label{tab:sim_user}
    \vspace{-10pt}
\end{table}

\begin{enumerate}
    \item \textbf{Uniform Curriculum} Cycles through all sub-skills in a fixed round-robin order regardless of learner proficiency. This represents a non-adaptive baseline that treats all learners identically and does not respond to individual learning patterns.

    \item \textbf{Greedy Curriculum} Reactively selects the sub-skill with lowest recent performance (highest completion time or failure rate) without maintaining an explicit proficiency model. This myopic strategy may over-practice already-mastered skills due to temporary performance fluctuations (slips) or under-practice skills that appear easy due to lucky guesses. It also uses a fixed exploration rate of 0.1.

    \item \textbf{Adaptive Curriculum (Ours)} Maintains explicit Bayesian proficiency estimates for each sub-skill and selects practice targets to maximize expected learning progress. The curriculum balances exploration (practicing uncertain skills) and exploitation (reinforcing near-mastery skills) using epsilon-greedy with a fixed exploration rate of 0.1~\cite{Wilson2016BackTT}.
\end{enumerate}

\noindent\textbf{Experimental Setup} For each curriculum strategy and sub-skill count (3, 5, 7, 10, 15), we generated 500 diverse simulated learners by sampling parameters from the distributions in Table~\ref{tab:sim_params}. By adding one additional sub-skill, we increase the allowable number of teaching actions by 50 and evaluate success as the proportion of learners achieving proficiency $\geq 0.8$ across all sub-skills.

\noindent\textbf{Results and Analysis} Table~\ref{tab:sim_user} shows success rates across curriculum complexity levels. The adaptive curriculum consistently outperforms both baselines across all complexity levels. For 15 sub-skills, adaptive achieves 93.6\% success rate compared to 66.0\% for greedy and 90.2\% for uniform. The performance gap widens as the number of sub-skills increases, demonstrating the importance of maintaining an explicit proficiency model for complex skill hierarchies.

The greedy strategy performs worst, particularly for complex curricula (15 sub-skills: 66.0\% success). Analysis of individual learner trajectories reveals two failure modes:
\begin{enumerate}
    \item \textbf{Thrashing}: Switching between sub-skills too frequently due to performance noise, preventing sustained practice on any single skill
    \item \textbf{Premature abandonment}: Moving away from near-mastery skills due to temporary slips, missing opportunities to consolidate learning
\end{enumerate}

\noindent\textbf{Uniform Curriculum Robustness} The uniform strategy performs surprisingly well (90.2\% for 15 sub-skills), nearly matching the adaptive approach. This suggests that for learners with sufficient practice budget, systematic coverage of all skills can be effective. However, uniform wastes resources on already-mastered skills and may frustrate learners with repetitive practice.

\noindent\textbf{Complexity Scaling} All strategies show declining success rates as sub-skill count increases, reflecting the fundamental challenge of mastering complex hierarchies within fixed practice budgets. The adaptive curriculum's advantage grows with complexity.

\noindent\textbf{Statistical Significance} We conducted pairwise t-tests comparing teaching actions used for each learner across curriculum strategies. Table~\ref{tab:sim_significance} shows p-values and Cohen's d effect sizes across all skill counts.

\begin{table}[!t]
    \centering
    \setlength{\tabcolsep}{6pt}
    \caption{Statistical significance of pairwise comparisons between curriculum strategies. P-values from t-tests and Cohen's d effect sizes for teaching actions used. Negative d values indicate Adaptive uses fewer actions.}
    \begin{tabular}{ccccc}
        \toprule
        \textbf{\#Skills} & \textbf{Comparison} & \textbf{p-value} & \textbf{Cohen's d} \\
        \hline
        \multirow{2}{*}{3} & Adaptive vs. Greedy & $<$0.0001 & $-$0.268 \\
                           & Adaptive vs. Uniform & $<$0.0001 & $-$0.332 \\
        \hline
        \multirow{2}{*}{5} & Adaptive vs. Greedy & $<$0.0001 & $-$0.404 \\
                           & Adaptive vs. Uniform & $<$0.0001 & $-$0.509 \\
        \hline
        \multirow{2}{*}{7} & Adaptive vs. Greedy & $<$0.0001 & $-$0.453 \\
                           & Adaptive vs. Uniform & $<$0.0001 & $-$0.636 \\
        \hline
        \multirow{2}{*}{10} & Adaptive vs. Greedy & $<$0.0001 & $-$0.664 \\
                            & Adaptive vs. Uniform & $<$0.0001 & $-$0.792 \\
        \hline
        \multirow{2}{*}{15} & Adaptive vs. Greedy & $<$0.0001 & $-$0.776 \\
                            & Adaptive vs. Uniform & $<$0.0001 & $-$0.781 \\
        \bottomrule
    \end{tabular}
    \label{tab:sim_significance}
    \vspace{-10pt}
\end{table}

\subsection{Study 1: Frame Analysis Evaluation Protocol}
\label{appendix:frame_analysis}

This section describes the detailed methodology for evaluating VLM frame analysis accuracy. Accurate frame analysis is critical for the coaching pipeline, as errors in state extraction propagate through timeline summarization and feedback generation.

\noindent\textbf{Dataset Construction} We randomly sample 100 frames from navigation task videos where users are either entering a door or using an elevator with the robot guide dog. Frame selection criteria:
\begin{itemize}
    \item \textbf{Temporal diversity}: Frames sampled from different time points (beginning, middle, end) of episodes to capture varying interaction states
    \item \textbf{Action diversity}: Frames showing different user actions (approaching, positioning, door manipulation, entry)
\end{itemize}

These frames are randomly assigned to two different human annotators who provide natural-language descriptions of:
\begin{itemize}
    \item User body position and posture (e.g., ``user standing 1.5m from door, facing slightly left'')
    \item Robot position relative to user (e.g., ``robot positioned to user's right, 0.5m ahead'')
    \item Environmental context (doorway, obstacles, spatial layout)
    \item User actions or errors visible in the frame (e.g., ``user reaching for handle but hand positioned too high'')
\end{itemize}

\noindent\textbf{Annotation Quality Control} To ensure annotation quality, we used 5 randomly selected samples as a quality control check. We manually evaluated these 5 annotations for consistency and accuracy against expert assessments. If the annotations showed inconsistencies or quality issues (e.g., missing critical details, factual errors, vague descriptions), we recruited a new annotator to re-annotate the entire dataset. 

\begin{figure}[t]
    \centering
    \begin{subfigure}[t]{0.32\textwidth}
        \vspace{0pt}
        \includegraphics[width=\textwidth]{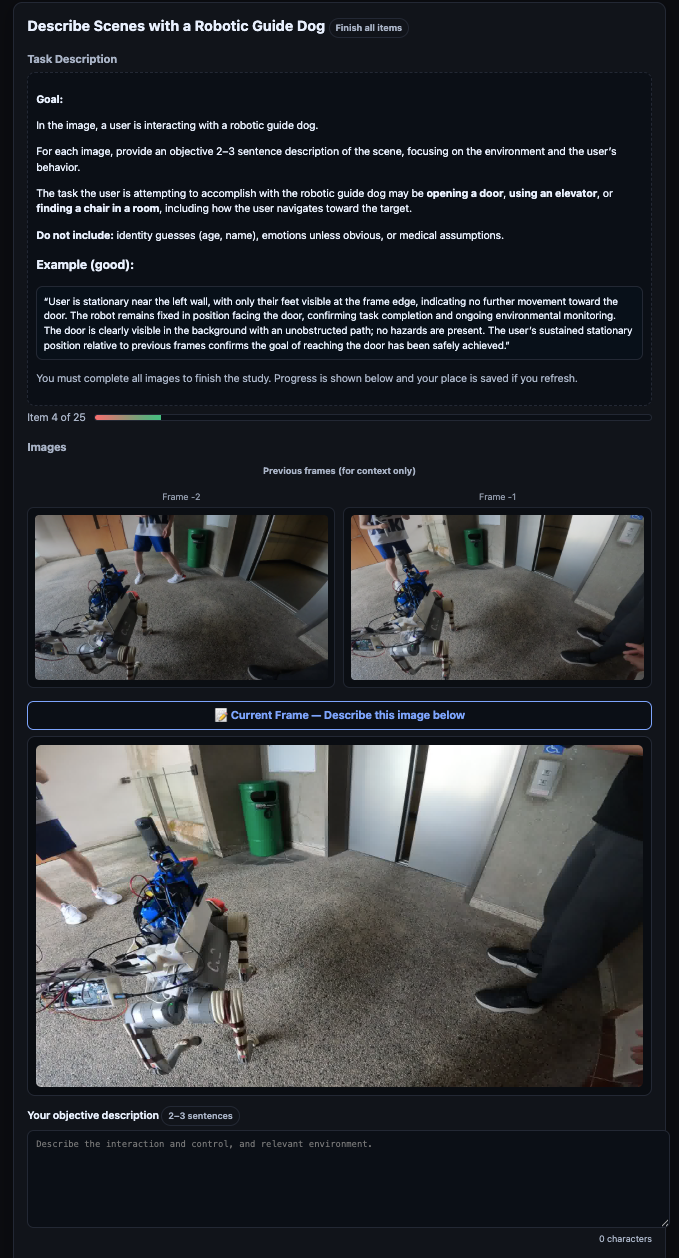}
        \caption{Labeling UI}
        \label{fig:human_label_ui}
    \end{subfigure}
    \hfill
    \begin{subfigure}[t]{0.32\textwidth}
        \vspace{0pt}
        \includegraphics[width=\textwidth]{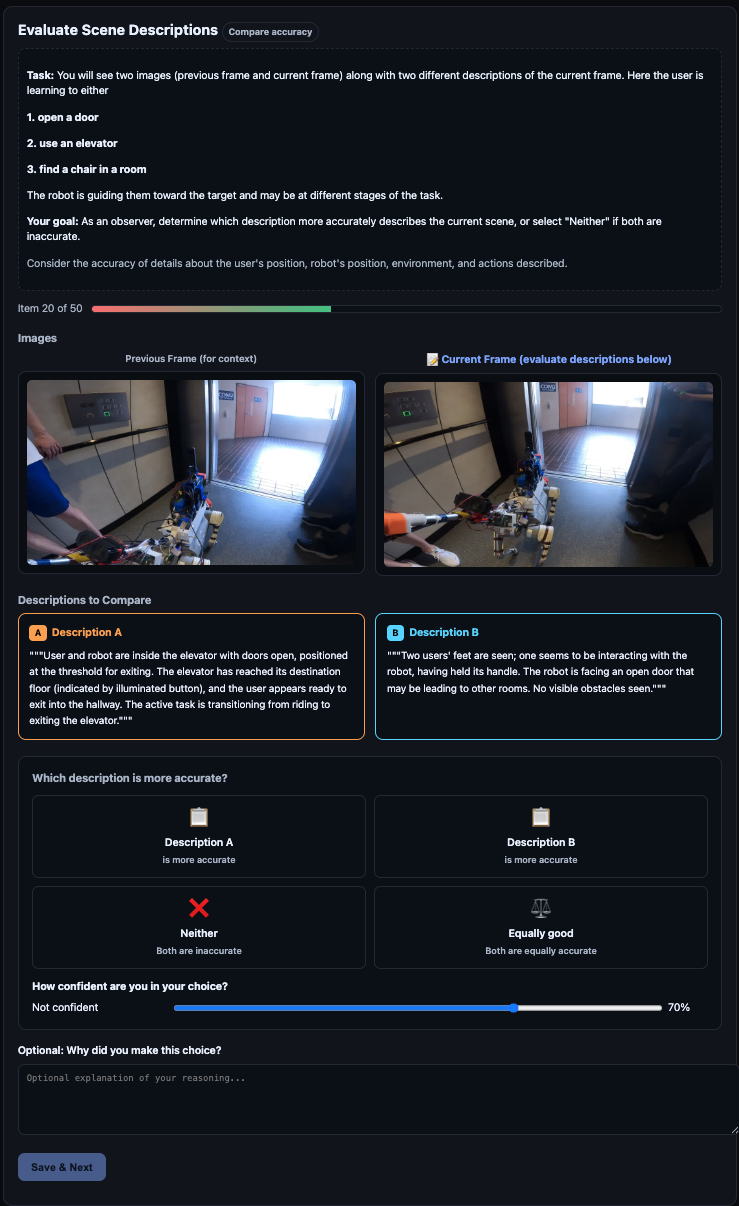}
        \caption{Frame Analysis UI}
        \label{fig:human_judge_ui}
    \end{subfigure}
    \hfill
    \begin{subfigure}[t]{0.32\textwidth}
        \vspace{0pt}
        \includegraphics[width=\textwidth]{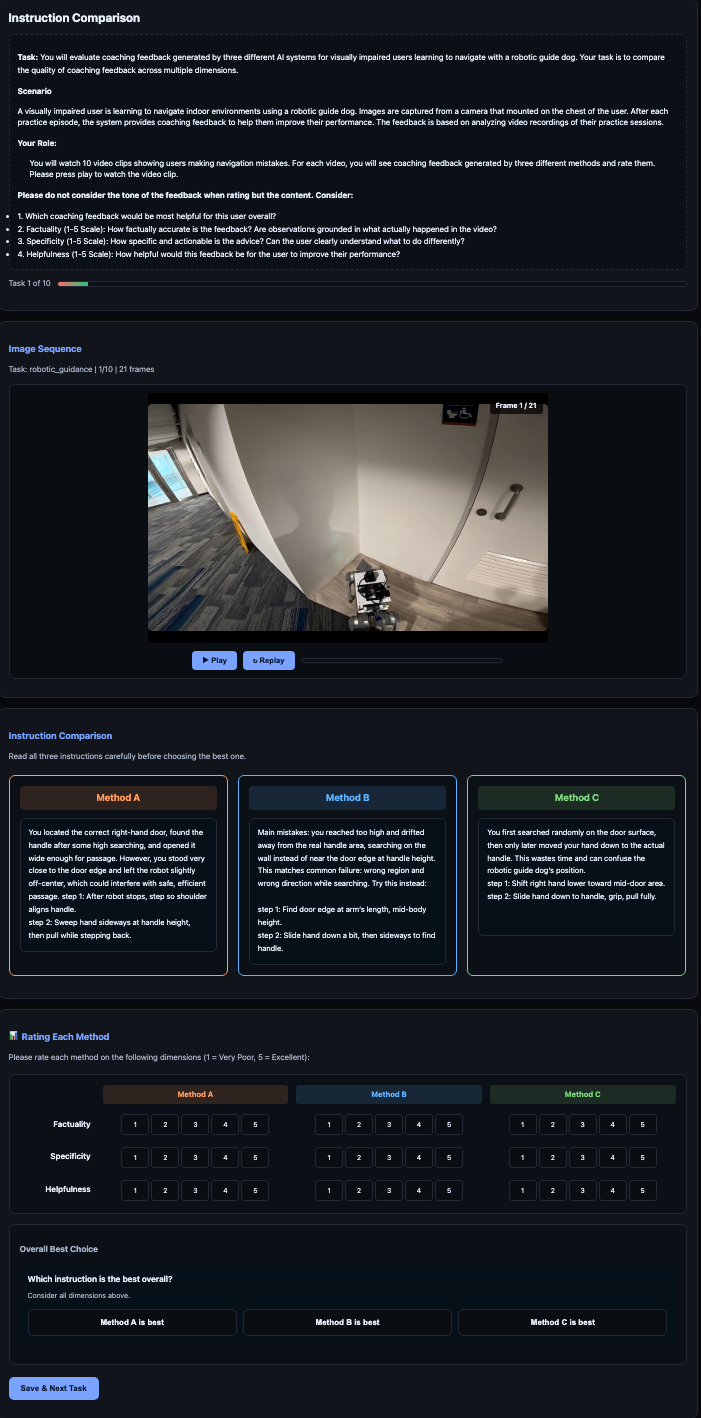}
        \caption{VLM Comparison UI}
        \label{fig:vlm_comp_ui}
    \end{subfigure}
    \caption{User interfaces used for data collection and evaluation. (a) Interface for human annotators to label ground truth frame descriptions. (b) Interface for judges to rate the faithfulness of VLM outputs against ground truth. (c) Interface for judges to compare coaching feedback from different models (Ours vs. Baselines).}
    \label{fig:all_uis}
\end{figure}

\noindent\textbf{Annotator Recruitment and Compensation} Annotators were recruited through Prolific with the following requirements:
\begin{itemize}
    \item Native English speakers
    \item Approval rate $\geq$ 95\% on previous Prolific tasks
    \item Passed the quality control checks
\end{itemize}
Annotators were compensated with 9 GBP per hour. It took approximately 1 hour to complete the 50-frame annotation task (each annotator annotated 50 frames). The UI for annotators can be found in~\figref{fig:all_uis}.

\noindent\textbf{VLM Prediction Generation} For each frame, we prompt the VLM (GPT-5.1) to generate structured observations following our frame analysis schema (detailed in Appendix~\ref{appendix:prompts}). The VLM receives:
\begin{itemize}
    \item The frame image
    \item Task context (``User is navigating with a robot guide dog to open a door'')
    \item Structured output schema specifying required fields (user state, robot state, environmental context, observed actions/errors)
    \item No access to human annotations (to ensure independent evaluation)
\end{itemize}

The prompt instructs the VLM to be factual and specific, avoiding speculation about user intent or internal states that are not directly observable.

\noindent\textbf{Human Evaluation Protocol} Human judges evaluate VLM prediction accuracy through pairwise comparisons. For each frame, judges are shown:
\begin{itemize}
    \item The original frame image
    \item Human annotation (Description A)
    \item VLM prediction (Description B)
\end{itemize}

Judges rate which description is more accurate using a multiple choice question:
\begin{itemize}
    \item A is more accurate
    \item B is more accurate
    \item Equally accurate
    \item Neither is accurate
\end{itemize}

\noindent\textbf{Evaluation Procedure}
\begin{enumerate}
    \item The order of A/B is randomized to prevent position bias
    \item Judges are instructed to evaluate based on the accuracy of the description. 
    \item For each model, we recruited 4 annotators to rate the pairwise comparison
    \item Each human judge evaluated 50 frames
    \item Judges were blind to which description came from human vs. VLM
\end{enumerate}
\noindent\textbf{Inter-Rater Agreement} Table~\ref{tab:frame_analysis_agreement} shows inter-rater agreement metrics, including Fleiss's kappa and unanimous agreement rates when judges had clear preferences (excluding ``equally accurate'' responses). 
\begin{table}[!t]
\centering
\caption{\textbf{Inter-rater agreement for frame analysis evaluation.} Four independent judges rated VLM-generated descriptions versus human annotations. Fleiss's kappa measures overall agreement. Unanimous agreement shows the percentage of frames. Both are for where judges had a clear preference for the same description (excluding ``equally accurate'' or ``neither accurate'' ratings).}
\label{tab:frame_analysis_agreement}
\begin{tabular}{lcc}
\toprule
\textbf{Model} & \textbf{Fleiss's Kappa} & \textbf{Unanimous Agreement (\%)} \\
\midrule
GPT-5.1 & 0.49 & 81.7\% \\
Claude-4.5-Sonnet & 0.55 & 84.9\% \\
Gemini-3.0-Pro & 0.32 & 83.9\% \\
Qwen3-VL & 0.44 & 82.5\% \\
\bottomrule
\end{tabular}
\vspace{-10pt}
\end{table}

\subsection{Study 1: VLM Comparison Methodology}
\label{appendix:vlm_comparison}

This section provides detailed methodology for comparing our decomposed four-stage pipeline against end-to-end VLM baselines. The comparison evaluates whether decomposing the coaching task into explicit stages (frame analysis, timeline summarization, feedback generation, robot adaptation) provides benefits over monolithic end-to-end processing.

We compare three approaches:

\noindent\textbf{Baseline 1: VLM (Direct Prompting)} The VLM receives the entire video (10-30 seconds, sampled at 2 fps) and a generic prompt requesting coaching feedback. This represents a minimal-effort baseline that relies entirely on the VLM's general capabilities without task-specific engineering.

\noindent\textbf{Baseline 2: Structured VLM} The VLM receives the entire video and structured prompt templates similar to our decomposed pipeline. The prompt includes:
\begin{itemize}
    \item The same schemas used in our pipeline (user state categories, common failure modes, actionable feedback templates)
    \item Explicit instructions to: (1) identify what happened, (2) diagnose why errors occurred, (3) provide specific actionable guidance
    \item Request for structured output matching our feedback format
\end{itemize}

However, unlike our pipeline, the VLM processes the entire video in a single call without decomposition into stages. This tests whether decomposition provides benefits beyond structured prompting alone.

\subsection{Study 2: Pipeline I/O Specification}
\label{appendix:pipeline_io}

\noindent\textbf{Ours (Decomposed Pipeline)} Our four-stage pipeline as described in \ifdefined\standaloneappendix the main paper\else Section~\ref{sec:method}\fi:
\begin{enumerate}
    \item Frame analysis ($f_{\text{frame}}$): Extract structured observations from individual frames
    \item Timeline summarization ($f_{\text{time}}$): Aggregate frame observations into episode-level summary
    \item Coaching generation ($f_{\text{coach}}$): Generate personalized feedback based on timeline and proficiency model
    \item Robot adaptation ($f_{\text{param}}$): Adjust robot parameters based on diagnosed errors
\end{enumerate}
Table~\ref{tab:pipeline_io} summarizes the input and output contract for each stage, making the intermediate representations used by the decomposed pipeline explicit.
\begin{table}[!t]
   \centering
   \setlength{\tabcolsep}{3pt}
   \caption{\redtext{\textbf{Pipeline inputs and outputs.} Each stage consumes structured outputs from the preceding stage, reducing open-ended generation and making the worked example in the supplementary material directly traceable.}}
   {\color{black}
   \begin{tabular}{p{0.22\columnwidth}p{0.30\columnwidth}p{0.34\columnwidth}}
   \toprule
   Stage & Input & Output \\
   \midrule
   $f_{\text{frame}}$ & Sampled RGB frames and task context & Per-frame state $h_t$: user, robot, environment, progress, safety flags \\
   $f_{\text{time}}$ & Frame states $h_{1:T}$ & Episode summary $\tau$: outcome, metrics, timestamped events \\
   $f_{\text{coach}}$ & Summary $\tau$ and transcribed user feedback & Coaching action $c$: diagnosis and terminal verbal instruction \\
   $f_{\text{param}}$ & Coaching action $c$ and proficiency belief & Robot adjustment $u$: conservative scaffolding parameters \\
   \bottomrule
   \end{tabular}}
   \label{tab:pipeline_io}
   \vspace{-7pt}
\end{table}

\noindent\textbf{Video Selection} We selected 10 navigation videos where users made clear errors requiring coaching intervention:
\begin{itemize}
    \item \textbf{Error types}: Wrong door handle search (3 videos), following too closely (3 videos), poor door coordination (3 videos), good performance (1 video)
    \item \textbf{Video length}: 10.5-31.5 seconds (mean 17.45s)
\end{itemize}

\noindent\textbf{Quality Control} To ensure reliable ratings, we used 2 videos as quality control checks:
\begin{enumerate}
    \item \textbf{Easy video}: Clear, obvious error with straightforward correction, user followed the robot too closely (expected high agreement)
    \item \textbf{Difficult video}: Subtle, ambiguous error requiring careful observation, user searched for the door handle for a long time on the wall (expected lower agreement but consistent with expert assessment)
\end{enumerate}

Each judge evaluated these control videos, and we manually verified their ratings against expert assessments. If a judge's ratings on both control videos showed significant discrepancies from the expert baseline (e.g., rated obviously incorrect feedback as helpful), we excluded that judge's ratings from the final analysis and recruited a replacement judge. Five judges were replaced using this procedure.

\noindent\textbf{Evaluation Metrics}

\begin{enumerate}
    \item \textbf{Preference Rate} Eight human judges picked their preferred generated feedback from the three approaches for each video. Judges were shown:
    \begin{itemize}
        \item The original video
        \item Three feedback texts (order randomized, approach labels hidden)
        \item Instructions to select the feedback that would be most helpful for learning
    \end{itemize}
    Preference rate is the percentage of videos where an approach received the most votes (plurality winner). 
    \item \textbf{Quality Ratings} In addition to preference rankings, judges rate each approach's feedback independently on four dimensions using 5-point Likert scales (1=poor, 5=excellent):
    \begin{itemize}
        \item \textbf{Faithfulness}: Accuracy of observations about what actually happened in the video (e.g., does the feedback correctly identify user actions and errors?)
        \item \textbf{Specificity}: Level of actionable detail provided (e.g., ``move hand forward 20cm'' vs. ``adjust hand position'')
        \item \textbf{Helpfulness}: Overall utility for learning and skill improvement (e.g., would this feedback help the user improve on the next attempt?)
    \end{itemize}
    \item \textbf{Latency} Time from video input to feedback generation, measured on identical hardware calling the same API endpoints. We report average time across all 10 videos.
\end{enumerate}

\noindent\textbf{Latency Breakdown.} Direct video prompting required 44.36\,s on average, structured full-video prompting required 42.15\,s, and our decomposed pipeline required 20.62\,s. For our pipeline, the dominant cost is the parallel frame-analysis stage; timeline summarization, coaching generation, and robot-parameter adaptation operate on compact structured text outputs. Because all stages run after an episode completes, this latency supports terminal feedback between attempts but not concurrent feedback during safety-critical navigation.

Judges rated all three approaches for each video, allowing within-video comparisons to control for video difficulty.

\paragraph{Inter-Rater Agreement} For preference rankings, we report agreement metrics appropriate for ranked data:
\begin{itemize}
    \item \textbf{Unanimous agreement}: All judges ranked the same method first (10.0\% of videos)
    \item \textbf{Majority agreement}: $\geq$50\% of judges agreed on top method (80.0\% of videos)
    \item \textbf{Average modal share}: Mean proportion of judges agreeing on the most-preferred method per video (58.8\%)
\end{itemize}

\subsection{Study 2: Detailed Experimental Protocol}
\label{appendix:protocol}

This section provides the full timeline and procedural details for Study 2 (human subject evaluation with blindfolded sighted participants).

Total session time is approximately 40 minutes. 

\begin{enumerate}
    \item \textbf{Introduction (5 min):}
    \begin{itemize}
        \item Consenting and briefing.
        \item Rigid leash fitting and adjustment; blindfold fitting; camera and microphone adjustment.
    \end{itemize}    
    \item \textbf{Practice Phase (7 min, 2 trials):}
    \begin{itemize}
        \item Two different routes that lead to different stopping points, distances, and facing directions from the door. 
        \item Purpose: Collect pretest performance to establish baseline.
    \end{itemize}
    
    \item \textbf{Training Phase (11 min):}
    \begin{itemize}
        \item Routes: Varying complexity, including different starting and ending positions, ending facing directions, and directional changes
        \item \emph{Baseline group:} 3 full end-to-end trials with task-level verbal instructions only (e.g., ``Now open the door'') without sub-skill decomposition, personalized error diagnosis, or adaptive curriculum sequencing
        \item \emph{\name{} group:} Up to 6 sub-skill episodes selected by the adaptive curriculum, with the session ending early if the system determined all sub-skills were mastered. Terminal feedback provided after each episode with personalized error diagnosis and adaptive curriculum sequencing. In our experiments, 4 participants mastered all sub-skills in less than 6 episodes.
        \item \emph{Safety timeout:} For each sub-skill training, a 25-second timeout was set for each sub-skill episode other than navigate to door to trigger human intervention if the participant was unable to complete the sub-skill within this time window.
    \end{itemize}
    
    \item \textbf{Evaluation Phase (7 min, 2 trials):}
    \begin{itemize}
        \item Different routes with similar complexity to training routes
        \item No feedback provided to assess skill learning
    \end{itemize}
    
    \item \textbf{Survey (10 min):} NASA-TLX, SUS, Customized Questionnaires for Perceived Usefulness and Perceived Ease of Use.
\end{enumerate}

\noindent\textbf{Route Length Compensation for Navigate to Door} When calculating skill improvement for the Navigate to Door sub-skill across different phases (practice, evaluation, retention), we compensate for differences in route length to ensure fair comparison. Since different routes have varying distances from the starting position to the door, raw completion times are not directly comparable.

We normalize performance using expert baseline times for each route configuration:
\begin{itemize}
    \item \textbf{Practice phase routes}: Expert baseline = 14.25s
    \item \textbf{Evaluation phase routes}: Expert baseline = 18.51s  
    \item \textbf{Retention phase routes}: Expert baseline = 18.91s
\end{itemize}

For each participant's Navigate to Door completion time $t_{\text{participant}}$ on a given route, we compute the normalized performance as:
\begin{equation}
t_{\text{normalized}} = t_{\text{participant}} - (t_{\text{expert}} - 14.25)
\end{equation}
where $t_{\text{expert}}$ is the expert baseline for that specific route. This adjustment removes the effect of route length differences, allowing us to measure genuine skill improvement rather than artifacts of route difficulty. All Navigate to Door performance comparisons reported in the main paper use these normalized times.

\noindent\textbf{Optimal Performance} For the knowledge tracing model, we define optimal performance times for each sub-skill based on expert demonstrations:
\begin{itemize}
    \item \textbf{Navigate to Door}: 15s (optimal expert time)
    \item \textbf{Open Door}: 5s (optimal expert time)
    \item \textbf{Enter Room}: 10s (optimal expert time)
\end{itemize}

For the novice level, we set all sub-skill performance estimates to 40s.

\subsection{Study 2: Baseline Instruction}
\label{appendix:baseline_instruction}

This section documents the task-level verbal instructions provided to the baseline group during Study 2. Unlike the \name{} group, which received personalized error diagnosis, adaptive curriculum sequencing, and sub-skill decomposition, the baseline group received only simple task-level instructions without detailed feedback.

\noindent\textbf{Initial Instruction (Before Practice Phase)}

Participants in both groups received the following initial briefing:
\begin{quote}
``I am a robot guide dog. I will help you navigate to the bathroom, pull the door, and go into the bathroom.''
\end{quote}

\noindent\textbf{Phase-Level Instructions During Training}

During the training phase, the baseline group received brief task-level prompts at the start of each phase, without personalized error diagnosis or corrective feedback:
\paragraph{Navigate to Door Phase}
\begin{quote}
``I'm leading you to the door.''
\end{quote}

\paragraph{Open Door Phase}
\begin{quote}
``You are arriving at the door. The handle is about 1.5 meters in front of you, on your right side. Please pull it open.''
\end{quote}

\paragraph{Enter Room Phase}
\begin{quote}
``The door is opened successfully. I'm leading you to the bathroom.''
\end{quote}

\subsection{Study 2: Coaching Feature Preference Survey}
\label{appendix:preference_survey}

This section provides the questionnaire used to assess which coaching features participants found most valuable in Study 2.

\noindent\textbf{Survey Instrument} After completing all trials, participants were asked to rank the following coaching features using a 7-point Likert scale from least helpful (1) to most helpful (7):

\begin{itemize}
    \item \textbf{Personalized Feedback}: The system provided feedback tailored to my specific mistakes and performance.
    \item \textbf{Sub-skill Practice}:  The ability to practice individual components.
    \item \textbf{Adaptive Curriculum}: The system automatically selected which navigation sub-skills to practice based on my proficiency.
    \item \textbf{Progressive Difficulty}: The system adjusted the navigation difficulty (e.g., distance to the door) based on my performance.
    \item \textbf{Terminal Feedback Timing}: Receiving feedback after each navigation trial rather than during navigation.
    \item \textbf{Feedback System}: I can provide feedback to the system.
\end{itemize}

Participants were also asked open-ended questions:
\begin{enumerate}
    \item Please rank your top 3 most valuable features from the list above 
    \item What aspects of the navigation task did you find most challenging?
    \item What aspects of the teaching method were most helpful for your learning?
    \item Did you develop any specific strategies for successful navigation? If so, please describe them.
    \item Were there any moments during navigation where you felt particularly unsafe or uncomfortable? If so, please describe them.
    \item Do you have any additional comments or suggestions about the robot guide dog system or the teaching method?
\end{enumerate}

Rankings were analyzed using mean rank scores (higher is better) and frequency of top-3 selections. The top-3 selections are 1) Personalized Feedback, 2) Adaptive Curriculum, and 3) Sub-skill Practice. 

\subsection{Study 2: Perceived Usefulness and Ease of Use Questionnaire}
\label{appendix:usefulness_questionnaire}

During the pilot study, we observed that participants had difficulty distinguishing between the coaching system component and the robot guide dog system itself. Since both the navigation guidance and the coaching feedback were delivered verbally, participants often conflated the two systems when providing feedback. For example, when asked about the usefulness of the system, participants would sometimes refer to the guide dog's navigation capabilities rather than the coaching feedback they received. To address this issue, we designed a customized questionnaire that explicitly asks participants to distinguish between two distinct components:

\noindent\textbf{Questionnaire Items} Table~\ref{tab:usefulness_questionnaire} presents the questionnaire items used to assess Perceived Usefulness and Perceived Ease of Use of the coaching system in Study 2, along with their respective Cronbach's alpha reliability coefficients. Participants rated each item on a 7-point Likert scale (1 = Strongly Disagree, 7 = Strongly Agree).

\begin{table}[t]
\centering
\caption{Perceived Usefulness and Perceived Ease of Use Questionnaire Items}
\label{tab:usefulness_questionnaire}
\begin{tabular}{p{0.85\linewidth}}
\toprule
\textbf{Perceived Usefulness} (Cronbach's $\alpha = 0.836$) \\
\midrule
1. Using the robot guide dog training system would improve my performance during robot guide dog handling tasks. \\
2. Using the robot guide dog training system would make robot guide dog training tasks easier to perform. \\
\midrule
\textbf{Perceived Ease of Use} (Cronbach's $\alpha = 0.859$) \\
\midrule
1. Learning to operate the robot guide dog training system would be easy for me. \\
2. My interaction with the robot guide dog training system would be clear and understandable. \\
3. It would be easy for me to become skillful at using the robot guide dog training system. \\
\bottomrule
\end{tabular}
\end{table}

\subsection{Study 2: Qualitative Feedback}
\label{appendix:qualitative_feedback}

This section summarizes additional qualitative feedback and system improvement suggestions from Study 2 participants. We conducted thematic analysis of open-ended survey responses and identified recurring themes.

\noindent\textbf{Positive Feedback}
\begin{enumerate}

\item \textbf{Personalized Error Diagnosis} Participants appreciated specific, actionable feedback tailored to their mistakes:
\begin{quote}
``The system answered my questions, cleared my confusion. It also suggested tips for how to improve in next trial. '' (P7, \name{} group)
\end{quote}

\begin{quote}
``The personalized feedback is accurate'' (P8, \name{} group)
\end{quote}

\item \textbf{Sub-Skill Decomposition} Breaking down the complex task into manageable components reduced cognitive load:
\begin{quote}
``Practicing where the door was with respect to the robot is helpful'' (P10, \name{} group)
\end{quote}
\end{enumerate}

However, some participants wanted concurrent feedback for certain sub-skills (see improvement suggestions below).

\noindent\textbf{Improvement Suggestions} Participants suggested several system enhancements:
\begin{enumerate}
    \item \textbf{Concurrent Feedback for Specific Sub-Skills}: Some participants (4/10) wanted real-time feedback during execution for certain sub-skills, particularly ``Open Door'':
    \begin{quote}
    ``For entering the room, terminal feedback was fine. But for finding the door handle, I want real-time guidance.'' (P8, \name{} group)
    \end{quote}
    \item \textbf{Tone of Feedback}: Some participants suggested the tone of feedback should be more positive:
    \begin{quote}
    ``The feedback could be less harsh'' (P4, \name{} group)
    \end{quote}
    \begin{quote}
    ``The coach sounds condescending'' (P10, \name{} group)
    \end{quote}
\end{enumerate}

\subsection{Study 3: Retention Analysis Protocol}
\label{appendix:retention_protocol}

Five participants from Study 2 (N=5 from \name{}) agreed to return 2 weeks after their initial training session.

\paragraph{Session Structure (30 min)}
\begin{enumerate}
    \item \textbf{Re-familiarization (5 min):}
    \begin{itemize}
        \item Brief reminder of rigid leash and robot guidance system
        \item No new instruction or coaching provided
    \end{itemize}
    
    \item \textbf{Practiced Routes (10 min, 2 trials):}
    \begin{itemize}
        \item Same routes from the evaluation phase of Study 2
        \item Purpose: Assess retention of learned skills
        \item No feedback provided
    \end{itemize}
    
    \item \textbf{Novel Routes (10 min, 2 trials):}
    \begin{itemize}
        \item New routes with similar complexity but different layouts
        \item Combine elements from practiced routes (turns and doorways) in new configurations
        \item Purpose: Assess transfer and generalization
        \item No feedback provided
    \end{itemize}
    
    \item \textbf{Brief Interview (5 min):}
    \begin{itemize}
        \item Qualitative feedback on what they remembered from training
        \item What felt easier or harder compared to the initial session
        \item Confidence levels navigating with the robot
    \end{itemize}
\end{enumerate}

\paragraph{Evaluation Metrics}
\begin{itemize}
    \item \textbf{Performance Maintenance}: Comparison of completion time and navigation efficiency between final training trial (from Study 2) and retention session
\end{itemize}
All participants demonstrated confident navigation with the robot guide dog. When asked about the skills they learned, they noted the importance of paying attention to the dog’s orientation and allowing sufficient space for the dog to adjust.

\subsection{Study 4: Visually Impaired User Case Study Protocol}
\label{appendix:vi_protocol}

This exploratory case study evaluates \name{} on both the doorway navigation task and robotic handover task with a totally blind participant (Male, 70, with no residual vision). The participant currently uses a white cane for daily navigation and has prior experience with robot guide dogs. Total session time is approximately 60 minutes for navigation followed by 30 minutes for handover.

\paragraph{Navigation Task Session (60 min)}

\begin{enumerate}
    \item \textbf{Introduction and Consent (10 min):}
    \begin{itemize}
        \item Informed consent and compensation explanation
        \item Demographic questionnaire
        \item System explanation and safety protocols
        \item Rigid leash fitting and familiarization with the robot guide dog
    \end{itemize}
    
    \item \textbf{Practice with Adaptive Coaching (35 min):}
    \begin{itemize}
        \item Full \name{} adaptive coaching system active
        \item Routes of varying complexity (same doorway navigation task as Study 2)
        \item System adapts curriculum across three sub-skills: Navigate to Door, Open Door, and Enter Room
        \item Terminal feedback provided after each sub-skill episode
        \item Evaluation trials to measure improvement 
        \item \bluetext{The trial structure matched Study 2: two initial practice trials, up to six coached sub-skill episodes, and two evaluation trials.}
    \end{itemize}
    
    \item \textbf{Semi-Structured Interview (15 min):}
    \begin{itemize}
        \item Coaching feature preference ranking (same instrument as Study 2)
        \item System usability and learning experience
        \item VI-specific design considerations (multimodal feedback, system control)
        \item Audio recorded with participant consent
    \end{itemize}
\end{enumerate}

\noindent\textbf{Results}
In the initial trial, the participant failed to locate the door handle and complete the task. After coaching, the participant successfully completed the task in the evaluation trial. \bluetext{For the ``Open Door'' sub-skill, completion time improved from 22.4\,s before coaching to 9.1\,s after coaching.} The attached video illustrates this progression.

\paragraph{Interview Topics}

The semi-structured interview covered the following key topics:

\begin{itemize}
    \item \textbf{Coaching Feature Preference}: Ranking of coaching features (sub-skill decomposition, personalized feedback, adaptive curriculum, error-specific diagnosis, actionable guidance) - same instrument as Study 2
    \item \textbf{Learning Experience}: Perceived effectiveness of coaching, most/least helpful feedback types, comparison to previous mobility training experiences
    \item \textbf{VI-Specific Design Needs}: 
    \begin{itemize}
        \item Multimodal feedback preferences (audio, haptic, other modalities)
        \item Need for system control (ability to pause, replay, skip feedback)
    \end{itemize}
    \item \textbf{Feedback Timing}: Preference for terminal (post-episode) vs. concurrent (real-time) feedback during navigation
    \item \textbf{Most Challenging Sub-skill}: Identification of which sub-skill (Navigate to Door, Open Door, Enter Room) was most difficult and why
\end{itemize}

\paragraph{Robotic Handover Task Session (30 min)}

Following the navigation session, the participant engaged with a robotic handover task to evaluate system extensibility (detailed setup in Appendix~\ref{appendix:handover_setup}):

\begin{enumerate}
    \item \textbf{Task Introduction and Familiarization (5 min):}
    \begin{itemize}
        \item Explanation of robotic handover task and safety protocols
        \item Haptic familiarization with robot gripper and workspace boundaries
        \item Demonstration of object transfer process
    \end{itemize}
    
    \item \textbf{Practice with Adaptive Coaching (20 min):}
    \begin{itemize}
        \item Receive objects of varying types from robotic arm (light, heavy, and large objects)
        \item \name{} coaching system active
        \item System adapts curriculum across three handover sub-skills based on object category
        \item Participant requested concurrent (real-time) feedback during execution to adjust hand position
        \item Metrics: Success rate, completion time, learning progression
    \end{itemize}
    
    \item \textbf{Cross-Task Comparison Interview (5 min):}
    \begin{itemize}
        \item Comparison of navigation vs. handover coaching experience
        \item Feedback timing preferences: terminal feedback for navigation vs. concurrent feedback for handover
        \item Task-specific coaching requirements and design considerations
    \end{itemize}
\end{enumerate}

\subsection{Study 4: Robotic Handover Task Setup}
\label{appendix:handover_setup}

This section details the experimental setup for the robotic handover task used in Study 4 to evaluate the extensibility of \name{} to a different robotic assistance domain.

\noindent\textbf{Robotic System}
\begin{itemize}
    \item \textbf{Manipulator}: Franka Research 3 robot with parallel gripper.
    \item \textbf{Workspace}: Robot positioned at table height (75 cm) to facilitate natural handover posture for seated users.
    \item \textbf{Safety Features}: Built-in collision detection, configurable force limiting (10N threshold), emergency stop button within participant reach.
    \item \textbf{Perception}: RGB-D camera (Intel RealSense L515) mounted overhead (1.5m height) providing top-down view of handover workspace for video observation and hand pose tracking.
\end{itemize}

\paragraph{Handover Objects}
Three object categories representing different manipulation challenges as shown in~\figref{fig:handover_objects}:
\begin{itemize}
    \item \textbf{Light objects}: Potato chips, plastic spoon, card - requires basic hand positioning and timing coordination.
    \item \textbf{Heavy objects}: A row of yogurt, water bottle, heavy box - requires sufficient grip strength and stability during transfer.
    \item \textbf{Irregular-shaped objects}: Sandwich box, large bag, kettle - requires proper hand configuration to accommodate object shape and size.
\end{itemize}

\begin{table*}
\centering
\begin{tabular}{ccccccccc}
\includegraphics[width=0.07\columnwidth]{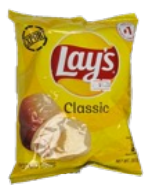} & 
\includegraphics[width=0.05\columnwidth]{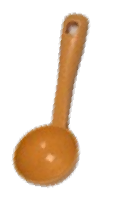} &
\includegraphics[width=0.07\columnwidth]{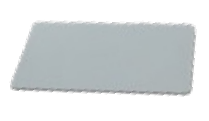} &
\includegraphics[width=0.11\columnwidth]{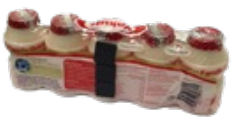} & 
\includegraphics[width=0.04\columnwidth]{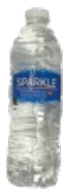} &
\includegraphics[width=0.11\columnwidth]{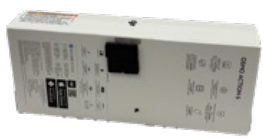} &
\includegraphics[width=0.09\columnwidth]{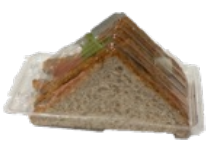} &
\includegraphics[width=0.08\columnwidth]{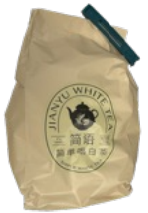} &
\includegraphics[width=0.12\columnwidth]{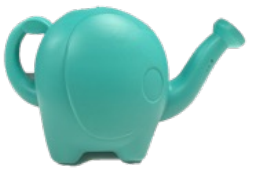} \\
Potato Chips & Plastic Spoon & Card & A Row of Yogurt & Water Bottle & Heavy Box & Sandwich Box & Large Bag & Kettle \\
\end{tabular}
\captionof{figure}{Objects used for handover experiment.}
\label{fig:handover_objects}
\end{table*}

\noindent\textbf{Task Structure and Sub-skill Decomposition.}In each trial, the robot moves from a home position to the handover location, transfers the object to the user's hand, and returns home. We decompose this interaction into three sub-skills based on object properties:

\begin{enumerate}
    \item \textbf{Handpose Alignment}: Put the hand in the correct position relative to the object.
    \begin{itemize}
        \item Focus: Hand positioning accuracy.
        \item Success criterion: Hand positioned appropriately before robot arrival.
        \item Common errors: Hand positioned too high/low/far/close.
    \end{itemize}
    \item \textbf{Light Object Handover}: Receiving light objects (100-200g).
    \begin{itemize}
        \item Focus: Hand positioning accuracy and timing coordination.
        \item Success criterion: Object transferred without dropping; hand positioned appropriately before robot arrival.
        \item Common errors: Hand positioned too high/low, premature or delayed grasp, insufficient contact area.
    \end{itemize}
    
    \item \textbf{Heavy Object Handover}: Receiving heavy objects (400-600g)
    \begin{itemize}
        \item Focus: Grip strength and weight preparation
        \item Success criterion: Object transferred without dropping; stable grip maintained during and after transfer
        \item Common errors: Insufficient grip force leading to drop, unprepared for object weight, grip released too early
    \end{itemize}
    
    \item \textbf{Irregular-shaped Object Handover}: Receiving irregular-shaped objects.
    \begin{itemize}
        \item Focus: Hand configuration and spatial awareness
        \item Success criterion: Object transferred without dropping; hand configuration accommodates object dimensions
        \item Common errors: Hand configuration too narrow for object, fingers not spread appropriately, unstable grasp due to poor hand shape
    \end{itemize}
\end{enumerate}

Each sub-skill requires mastering coordination without visual feedback, making personalized coaching essential for error diagnosis and skill improvement.

\noindent\textbf{Observation and State Representation}

In addition to the chest-mounted camera, the system also uses the overhead RGB-D camera as input to get the user's hand state: position relative to the handover location, hand orientation (palm facing direction), and hand configuration (open/closed, finger spread).

\noindent\textbf{Performance Metrics}
\begin{itemize}
    \item \textbf{Success Rate}: Proportion of trials where object was successfully transferred without dropping (primary outcome)
    \item \textbf{Completion Time}: Duration from robot departure from home position to successful stable grasp by user
    \item \textbf{Sub-skill Proficiency}: Success rate tracked separately for each object category (light, heavy, large) to inform adaptive curriculum
    \item \textbf{Error Distribution}: Frequency of different error types (positioning, timing, grip strength, hand configuration) to characterize learning challenges
\end{itemize}

\noindent\textbf{Results}
In the first three trials, the user was only able to receive the heavy object passively. After training with \name{}, the user learned to actively open the palm and grasp the object from the bottom. In the final four evaluation trials, the user successfully grasped all objects, including a challenging case that required grasping a kettle. The attached video illustrates this progression.

\noindent\textbf{Comparison to Navigation Task}

The handover task provides a complementary test case to evaluate \name{}'s generalizability:

\begin{itemize}
    \item \textbf{Interaction modality}: Fine-motor proprioceptive control (hand positioning, grip) vs. gross-motor spatial coordination (body movement, leash following)
    \item \textbf{Error manifestation}: Immediate binary failure (object drops) vs. accumulated inefficiency (longer completion time, suboptimal path)
    \item \textbf{Sensory demands}: Tactile and force feedback vs. auditory cues and spatial reasoning
\end{itemize}

Despite these fundamental differences, both tasks share the core coaching challenge: helping users master coordinated interaction with robotic systems without visual feedback. This allows us to evaluate whether \name{}'s architecture (adaptive curriculum, sub-skill decomposition, personalized feedback) generalizes beyond a single task domain.

\subsection{System Failure Modes}
\label{appendix:failure_modes}

We document observed system failures:

\begin{enumerate}
    \item \textbf{Failure Mode 1: Visibility Failure} When the user's body occludes the camera or the user changes the orientation entirely, the VLM cannot extract accurate state information.
    \item \textbf{Failure Mode 2: Lighting Conditions} Poor lighting (too dark or strong backlighting) degrades VLM performance.
    \item \textbf{Failure Mode 3: Motion Blur}  Rapid user movements create motion blur, making frame analysis unreliable.
    \item \textbf{Failure Mode 4: Contradictory Feedback} LLM generates feedback that contradicts previous feedback.
    \item \textbf{Failure Mode 5: Overly Complex Language} Feedback uses technical jargon or complex sentence structures difficult for users to understand.
    \item \textbf{Failure Mode 6: Wrong-format Feedback} The LLM failed to return a parsable JSON object.
    \item \textbf{Failure Mode 7: Communication Latency} Network latency causes delays in feedback delivery.
\end{enumerate}

\noindent\textbf{Lessons for Deployment} Key insights from failure mode analysis:
\begin{enumerate}
    \item \textbf{History is critical}: Multiple historical frames are necessary for accurate state estimation
    \item \textbf{Conservative defaults}: When uncertain, provide generic feedback rather than potentially incorrect specific feedback
    \item \textbf{Human oversight}: Maintain human supervision for safety-critical applications
    \item \textbf{Graceful degradation}: System should continue functioning (with reduced quality) rather than failing completely
    \item \textbf{User agency}: Allow users to override system decisions and request clarification
\end{enumerate}

\subsection{Prompt Templates}
\label{appendix:prompts}
\paragraph{VLM Baseline}
\stepcounter{prompt}
\begin{tcolorbox}[
    colback=gray!5,
    colframe=black,
    boxrule=0.5pt,
    arc=2pt,
    left=4pt,
    right=4pt,
    top=4pt,
    bottom=4pt,
    breakable,
    title=\textbf{(\theprompt) VLM Analysis},
    fonttitle=\small\bfseries,
]

Task: [Task Description according to the task]

You are analyzing a sequence of images from a robotic task. Please examine all the images provided and generate clear, actionable instructions to help complete the task. Please format your response as clear, concise instructions that can be followed. 

Please put your response in the following JSON format:

\begin{lstlisting}[breaklines=true,basicstyle=\ttfamily\footnotesize]
{    
    "terminal_instruction_to_user": "string - terminal instruction to the user to help them improve their navigation technique. First, point out mistakes if the performance is bad. Then, provide corresponding actionable suggestions in relative term. If possible, step by step, start with 'step 1', 'step 2', maximum two steps. For each step, be each point very concise, less than 10 words, no need to be a full sentence. In the end, answer the human user feedback in a very concise way. if there is no feedback, there is no need to answer the human user feedback.",
}
\end{lstlisting}
\end{tcolorbox}

\paragraph{Structured VLM}
\begin{tcolorbox}[
    colback=gray!5,
    colframe=black,
    boxrule=0.5pt,
    arc=2pt,
    left=4pt,
    right=4pt,
    top=4pt,
    bottom=4pt,
    breakable,
    title=\textbf{(\theprompt) Structured VLM Analysis},
    fonttitle=\small\bfseries,
]

Prompt: You are analyzing a camera image mounted on the human chest for robot guide dog navigation training. \\

\textbf{Inputs Provided:}

\textbf{Camera Image:} Single current frame from the camera mounted on the human chest showing:
\begin{itemize}
   \item User's body position and walking path
   \item Distance and positioning relative to robot
   \item Environment and navigation obstacles
   \item User's following behavior and spatial awareness
\end{itemize}

\textbf{Analysis Approach:}
\begin{itemize}
   \item Always pay attention to Human's hands behavior, e.g. if the human current pose can reach the door handle or not
   \item Distance between human and the door and door handle
   \item User difficulties (hesitation, confusion, searching, getting stuck)
   \item User mistakes that may leads to task failure or inefficient task completion
\end{itemize}

Be objective and specific. Base your analysis solely on visual observation of the camera image. \\

\textbf{Task Information:} [Task Description according to the sub-skill] \\

\textbf{Common Failure Modes to Watch For:}
\begin{lstlisting}[breaklines=true,basicstyle=\ttfamily\footnotesize]
{
  "1": "User search the wrong region for handles and door",
  "2": "User stands too far away from the door",
  "3": "User search the wrong region for handles and door in the wrong direction",
  "4": "User does not keep the door fully open for the robot to pass through",
  "5": "User is too close to the robot when passing through the door so when the robot adjust the pose, user may interfere the robot's adjustment",
}
\end{lstlisting}

\textbf{Instruction Generation:} \\

Based on ALL the images you see, please provide:
\begin{enumerate}
\item A summary of what is happening in the image sequence (chronological analysis)
\item Specific step-by-step instructions for what actions should be taken
\item Any important observations or warnings about the current state
\item Identification of any failure modes or potential issues observed
\item Maximum 2 steps and be very concise.
\end{enumerate}

Please format your response as clear, concise instructions that can be followed.
Focus on providing actionable guidance based on the complete sequence of images.
Please put your response in the following JSON format:
\begin{lstlisting}[breaklines=true,basicstyle=\ttfamily\footnotesize]
{    
    "technique_assessment": {
        "following_technique": "enum - excellent|good|needs_improvement|poor",
        "distance_maintenance": "enum - excellent|good|needs_improvement|poor",
        "cue_response": "enum - excellent|good|needs_improvement|poor",
        "trust_level": "enum - excellent|good|needs_improvement|poor"
    },
    "specific_feedback": {
        "strengths": ["list of things done well"],
        "areas_for_improvement": ["list of navigation technique improvements"],
        "safety_notes": ["list of safety observations during navigation"],
        "communication_feedback": ["specific feedback about robot communication"]
    },
    "coaching_recommendations": {
        "immediate_practice": ["navigation skills to practice in next session"],
        "technique_tips": ["specific navigation technique improvements"],
        "communication_tips": ["how to better notice and respond to robot signals"],
        "safety_tips": ["safety practices for navigation"]
    },
    "result_summary": "Brief navigation coaching summary",
    "terminal_instruction_to_user": "string - terminal instruction to the user to help them improve their navigation technique. First, point out mistakes if the performance is bad. Then, provide corresponding actionable suggestions in relative term. If possible, step by step, start with 'step 1', 'step 2', maximum two steps. For each step, be each point very concise, less than 10 words, no need to be a full sentence. In the end, answer the human user feedback in a very concise way. if there is no feedback, there is no need to answer the human user feedback.",
    "user_actionables": ["actionable navigation suggestion 1", "actionable navigation suggestion 2", "actionable navigation suggestion 3"]
}
\end{lstlisting}
\end{tcolorbox}

\paragraph{\name{}}

\stepcounter{prompt}

\begin{tcolorbox}[
    colback=gray!5,
    colframe=black,
    boxrule=0.5pt,
    arc=2pt,
    left=4pt,
    right=4pt,
    top=4pt,
    bottom=4pt,
    breakable,
    title=\textbf{(\theprompt) Frame Analysis},
    fonttitle=\small\bfseries,
]

You are analyzing a camera image mounted on the human chest for robot guide dog navigation training.

\textbf{Inputs Provided:}
1. Camera Image: Single current frame from the camera mounted on the human chest showing:
\begin{itemize}
    \item User's body position and walking path
    \item Distance and positioning relative to robot
    \item Environment and navigation obstacles
    \item User's following behavior and spatial awareness
\end{itemize}

\textbf{Analysis Approach:}
\begin{itemize}
    \item Always pay attention to Human's hands behavior, e.g. if the human current pose can reach the door handle or not
    \item Distance between human and the door and door handle
    \item User difficulties (hesitation, confusion, searching, getting stuck)
    \item User mistakes that may leads to task failure or inefficient task completion
\end{itemize}

Be objective and specific. Base your analysis solely on visual observation of the camera image.

Task: 
\begin{lstlisting}[breaklines=true,basicstyle=\ttfamily\footnotesize]
[Task Description according to the sub-skill]
\end{lstlisting}

\textbf{Analysis Inputs:}
\begin{itemize}
    \item Image: Current camera view from human chest-mounted camera showing the navigation scene
\end{itemize}

\textbf{Analysis Task:}
\begin{enumerate}
    \item What does the image show about user position and navigation?
    \item Is the user correctly positioned relative to the robot (if visible)?
    \item What is the navigation context and progress?
    \item What specific observations can be made about the navigation scenario?
    \item Is there adequate clearance and space for safe passage?
\end{enumerate}

Required per-frame schema:
\begin{lstlisting}[breaklines=true,basicstyle=\ttfamily\footnotesize]
{
    "timestamp_sec": "<float>",
    "user_state": "enum - unknown|approaching|positioned|interacting|completed|stuck|retreating|hesitant|confused|searching",
    "robot_state": "enum - unknown|guiding|waiting|signaling|assisting|monitoring|stopped|adjusting",
    "connection_state": "enum - connected|loose|disconnected|unknown - status of leash/harness connection between user and robot",
    "environment_state": "enum - unknown|clear|cluttered|hazardous|changing",
    "task_progress": "enum - not_started|in_progress|nearly_complete|complete|failed|interrupted",
    "safety_concerns": "boolean - are there any safety issues present",
    "user_difficulties": "boolean - is the user experiencing any difficulties or challenges",
    "frame_description": "<string> - a brief description of the frame, noting any challenges, safety issues, or user difficulties"
}
\end{lstlisting}
\end{tcolorbox}

\begin{tcolorbox}[
    colback=gray!5,
    colframe=black,
    boxrule=0.5pt,
    arc=2pt,
    left=4pt,
    right=4pt,
    top=4pt,
    bottom=4pt,
    breakable,
    title=\textbf{(\theprompt) Summarization Analysis},
    fonttitle=\small\bfseries,
]

You are analyzing a robot guide dog navigation episode to determine if the user successfully completed the navigation task.

\textbf{Success Criteria:}
\begin{enumerate}
    \item User followed the robot guide dog's lead and cues
    \item Proper distance maintained from the robot (about 1-2 feet)
    \item User responded appropriately to robot's directional commands
    \item Safe navigation without incidents or obstacles
    \item Task objective was achieved completely and safely (e.g., reached door, opened door, entered room)
    \item All actions were completed with adequate clearance and space for safe passage
\end{enumerate}

\textbf{Failure Indicators:}
\begin{enumerate}
    \item User pulled ahead of robot instead of following
    \item Incorrect distance maintained (too close or too far)
    \item User failed to respond to robot's directional cues
    \item User lost contact with robot's harness or guiding mechanism
    \item User lacked awareness of surroundings
    \item Task objective not achieved or completed incompletely
    \item Insufficient clearance or space created, potentially causing safety issues or blocking passage
\end{enumerate}

Task Information: 
\begin{lstlisting}[breaklines=true,basicstyle=\ttfamily\footnotesize]
[Task Description according to the sub-skill]
\end{lstlisting}

Timeline: 
\begin{lstlisting}[breaklines=true,basicstyle=\ttfamily\footnotesize]
[timeline_json]
\end{lstlisting}

Determine if what is the main mistake the user made across the timeline and write in the timeline consistently, e.g. too far from the door, wrong search region, dangerous behavior, etc.

Required decision schema:
\begin{lstlisting}[breaklines=true,basicstyle=\ttfamily\footnotesize]
{    
    "success": "boolean - did user successfully complete the navigation task",
    "navigation_quality": "enum - excellent|good|fair|poor|very_poor",
    "time_efficiency": "float - actual_time / target_time ratio",
    "safety_rating": "enum - excellent|good|fair|poor|very_poor",
    "following_technique": "enum - excellent|good|fair|poor|very_poor",
    "communication_effectiveness": "boolean - was user-robot communication clear",
    "trust_level": "enum - excellent|good|fair|poor|very_poor",
    "issues_encountered": ["list of any problems"],
    "result_summary": "Brief outcome description",
    "timeline_summary": "string - detailed chronological summary of what the user did at key timestamps, explicitly stating the timing and actions (e.g., 'At 0.5s: user established connection with guide dog. At 2.1s: user followed left turn cue. At 5.3s: user maintained proper distance. At 8.7s: user arrived at door.'). This summary will be used as input for coaching instruction generation."
}
\end{lstlisting}
\end{tcolorbox}

\stepcounter{prompt}
\begin{tcolorbox}[
    colback=gray!5,
    colframe=black,
    boxrule=0.5pt,
    arc=2pt,
    left=4pt,
    right=4pt,
    top=4pt,
    bottom=4pt,
    breakable,
    title=\textbf{(\theprompt) Coaching Generation},
    fonttitle=\small\bfseries,
]
You are a mobility instructor providing coaching feedback for a visually impaired person learning to use a robot guide dog.
Please be very critical and strict. Please pay attention to every detail.

Decision: [summarization\_json]\\
Task Info: [task\_info\_json] \\

Provide clear, easy-to-understand coaching for robot guide dog navigation skills. Write as if speaking directly to the user in a friendly, supportive way. Use simple words, short sentences, and focus on what they can feel or hear.

When providing guidance, consider the importance of completing actions with adequate clearance and space. Help users understand how to feel for full range of motion in their interactions, ensuring smooth and safe navigation.

\textbf{Common Failure Modes to Watch For:}
\begin{lstlisting}[breaklines=true,basicstyle=\ttfamily\footnotesize]
{
  "1": "User search the wrong region for handles and door",
  "2": "User stands too far away from the door",
  "3": "User search the wrong region for handles and door in the wrong direction",
  "4": "User does not keep the door fully open for the robot to pass through",
  "5": "User is too close to the robot when passing through the door so when the robot adjust the pose, user may interfere the robot's adjustment",
}
\end{lstlisting}

Required coaching schema:
\begin{lstlisting}[breaklines=true,basicstyle=\ttfamily\footnotesize]
{
    "technique_assessment": {
        "following_technique": "enum - excellent|good|needs_improvement|poor",
        "distance_maintenance": "enum - excellent|good|needs_improvement|poor",
        "cue_response": "enum - excellent|good|needs_improvement|poor",
        "trust_level": "enum - excellent|good|needs_improvement|poor"
    },
    "specific_feedback": {
        "strengths": ["list of things done well"],
        "areas_for_improvement": ["list of navigation technique improvements"],
        "safety_notes": ["list of safety observations during navigation"],
        "communication_feedback": ["specific feedback about robot communication"]
    },
    "coaching_recommendations": {
        "immediate_practice": ["navigation skills to practice in next session"],
        "technique_tips": ["specific navigation technique improvements"],
        "communication_tips": ["how to better notice and respond to robot signals"],
        "safety_tips": ["safety practices for navigation"]
    },
    "result_summary": "Brief navigation coaching summary",
    "terminal_instruction_to_user": "string - terminal instruction to the user to help them improve their navigation technique. First, point out mistakes if the performance is bad. Then, provide corresponding actionable suggestions in relative term. If possible, step by step, start with 'step 1', 'step 2', maximum two steps. For each step, be each point very concise, less than 10 words, no need to be a full sentence. In the end, answer the human user feedback in a very concise way. if there is no feedback, there is no need to answer the human user feedback.",
    "user_actionables": ["actionable navigation suggestion 1", "actionable navigation suggestion 2", "actionable navigation suggestion 3"]
}
\end{lstlisting}
\end{tcolorbox}

\stepcounter{prompt}
\begin{tcolorbox}[
    colback=gray!5,
    colframe=black,
    boxrule=0.5pt,
    arc=2pt,
    left=4pt,
    right=4pt,
    top=4pt,
    bottom=4pt,
    breakable,
    title=\textbf{(\theprompt) Parameters Tuning},
    fonttitle=\small\bfseries,
]
You are a robot parameter tuning engineer specializing in robot guide dog navigation systems.

For guide dog navigation tasks, focus on parameters that affect the safety and effectiveness of navigation guidance:
\begin{itemize}
    \item Distance to door: How close the robot maintains to the target (door)
    \item Teaching method: Whether to use scaffolding (more support) or challenging (less support) approach
\end{itemize}

Coaching Analysis Results:
\begin{lstlisting}[breaklines=true,basicstyle=\ttfamily\footnotesize]
{coaching_json}
\end{lstlisting}

User Feedback (Natural Language):
\begin{lstlisting}[breaklines=true,basicstyle=\ttfamily\footnotesize]
{user_feedback_json}
\end{lstlisting}

\textbf{Navigation-Specific Analysis:}
For robot guide dog navigation tasks, prioritize safety and effective following during navigation. \\

\textbf{Step 1 - Extract Navigation Feedback:}
Parse the user's natural language feedback for navigation-specific concerns:
\begin{itemize}
    \item Distance comfort: ``too close'', ``too far'', ``comfortable distance''
    \item Guidance clarity: ``confusing'', ``clear'', ``helpful''
\end{itemize}

\textbf{Step 2 - Analyze Navigation Technique:}
\begin{itemize}
    \item User's following behavior and positioning
    \item Safety considerations during navigation
\end{itemize}

\textbf{Step 3 - Recommend Navigation Parameter Adjustments:}
Based on the navigation analysis, recommend adjustments to:
\begin{lstlisting}[breaklines=true,basicstyle=\ttfamily\footnotesize]
    "distance_to_door": closer/farther/maintain (distance robot maintains to door)
    "pointing_direction": more_left/more_right/maintain (how the robot's pointing direction shifts horizontally: left, center, or right relative to the user)
\end{lstlisting}    

Required navigation parameter tuning schema:
\begin{lstlisting}[breaklines=true,basicstyle=\ttfamily\footnotesize]
{
    "distance_to_door_adjustment": "enum - closer|farther|maintain - adjust distance robot maintains to door",
    "pointing_direction_adjustment": "enum - more_left|more_right|maintain - adjust horizontal pointing direction (left/center/right)",
    "reasoning": "string - explanation for navigation parameter adjustments based on user safety and comfort"
} 
\end{lstlisting}
\end{tcolorbox}

\subsection{Formative Study Semi-Structured Interview Guide}
\label{appendix:formative_questionnaire}

We used the following 25-question semi-structured interview guide in the formative study. The interviewer asked open-ended questions, allowed participants to elaborate freely, and used follow-up prompts only to clarify participant-raised points. The guide was designed to elicit participants' own experiences rather than ask them to choose from predefined answers.

\noindent\textbf{Opening Script.}
``We are interested in your experience learning to use a robot guide dog and in what kinds of support would make that learning process easier or safer. There are no right or wrong answers. Please describe your experience in your own words, and feel free to skip any question.''

\subsubsection{Background and Prior Training}
\begin{enumerate}[leftmargin=*]
    \item What navigation aids or assistive technologies do you currently use in daily life?
    \item How did you learn to use your current navigation aid or assistive technology?
    \item What parts of that learning process were most helpful?
    \item What parts of that learning process were difficult or insufficient?
\end{enumerate}

\subsubsection{Robot Guide Dog Experience}
\begin{enumerate}[resume,leftmargin=*]
    \item Please describe your experience navigating with the robot guide dog. What are the most difficult tasks? 
    \item What aspects of the robot guide dog were easiest to understand?
    \item What aspects of the robot guide dog were confusing, difficult, or unexpected?
    \item How did you interpret the robot's motion, leash cues, or other signals?
    \item Did your behavior or understanding change over the practice session? If so, what changed?
\end{enumerate}

\subsubsection{Coaching and Feedback}
\begin{enumerate}[resume,leftmargin=*]
    \item Please describe the coaching or feedback you received during practice.
    \item Which feedback, if any, helped you change what you did on the next attempt?
    \item Were there moments when feedback would have been useful but was missing?
    \item When you made a mistake or felt uncertain, what information would have helped you recover?
    \item How should feedback be timed for this kind of task?
\end{enumerate}

\subsubsection{Practice Structure and Personalization}
\begin{enumerate}[resume,leftmargin=*]
    \item How would you prefer to practice a complex navigation task with a robot guide dog?
    \item Are there parts of the task that should be practiced separately? If so, which parts and why?
    \item How should practice difficulty change as a learner improves?
    \item What should a coaching system remember about an individual learner's progress?
\end{enumerate}

\subsubsection{Safety, Trust, and Sensing}
\begin{enumerate}[resume,leftmargin=*]
    \item Please describe any moments when you felt especially safe, unsafe, confident, or uncertain.
    \item What would increase your trust in the robot guide dog during training?
    \item Were there moments when the coach or robot seemed to lack information about what you were doing?
    \item What kinds of cues or interaction controls would help you manage cognitive load during training?
\end{enumerate}

\subsubsection{Future Automated Coaching}
\begin{enumerate}[resume,leftmargin=*]
    \item What would make an automated coaching system useful for learning to use robotic assistants?
    \item What should an automated coaching system avoid doing?
    \item Is there anything else about learning to use a robot guide dog or robotic assistant that we have not asked about?
\end{enumerate}

\end{document}